\pdfoutput=1
\documentclass{article}
\usepackage[margin=1in]{geometry}
\usepackage{eqparbox}

\usepackage{amsmath,amsfonts,bm}

\def\eqref#1{equation~\ref{#1}}
\def\1{\bm{1}}

\DeclareMathAlphabet{\mathsfit}{\encodingdefault}{\sfdefault}{m}{sl}
\SetMathAlphabet{\mathsfit}{bold}{\encodingdefault}{\sfdefault}{bx}{n}

\newcommand{\sigmoid}{\sigma}

\newcommand{\Pobs}{\mathbb{P}_{\mathrm{obs}}}
\newcommand{\Ptar}{\mathbb{P}_{\mathrm{tar}}}

\usepackage{url}

\usepackage{amsmath}
\usepackage{natbib}
\usepackage{enumitem}
\usepackage{graphicx}
\usepackage{mathtools}

\usepackage[dvipsnames]{xcolor}
\usepackage{tcolorbox}
\definecolor{Red}{rgb}{0.6,0,0}

\definecolor{policy1}{HTML}{76a5b0}
\definecolor{policy2}{HTML}{0c5394}

\usepackage[backref=page]{hyperref}
\hypersetup{colorlinks=true}
\hypersetup{linktoc=all}
\hypersetup{citecolor=MidnightBlue}
\hypersetup{linkcolor=Red}
\hypersetup{urlcolor=MidnightBlue}
\usepackage[all]{hypcap}
\usepackage[noabbrev, nameinlink, capitalize]{cleveref}
\usepackage{xcolor}
\usepackage{adjustbox}
\usepackage{multirow}
\usepackage{booktabs}
\usepackage{subcaption}
\usepackage{wrapfig}
\usepackage{graphicx}

\usepackage{etoc}
\usepackage{xcolor}

\def\ci{\perp\!\!\!\perp}
\newtheorem{assumption}{Assumption}[section]

\newtheorem{remark}{Remark}[section]

\title{Evaluation of Active Feature Acquisition Policies with \\ Tabular Foundation Models}

\author{%
  Yuta Kobayashi\thanks{Department of Biomedical Informatics, Columbia University. Correspondence to \texttt{yk3043@cumc.columbia.edu}}
  \and Divyam Madaan\footnotemark[1]
  \and Shalmali Joshi\footnotemark[1]
}

\date{}
\begin{document}

\maketitle

\begin{abstract}
Active feature acquisition learns policies that sequentially acquire features to maximize information about a target variable. We study how to
learn and evaluate such policies from finite offline data using prior-data fitted networks (PFNs), which are off-the-shelf models that output posterior predictive distributions without task-specific training. We show that under the imbalanced coverage of offline data, using total predictive entropy as a reward creates an epistemic bias that penalizes acquiring sparsely observed features. Specifically, this reward conflates \emph{epistemic uncertainty} (arising from lack of offline data) with \emph{aleatoric uncertainty} (arising from uninformative features). To address this, we target the posterior expected (aleatoric) entropy instead of the total predictive entropy output by a PFN for evaluating feature acquisitions. Empirical evaluations on synthetic and real-world datasets demonstrate that our approach consistently reduces value estimation bias and yields credible intervals with strong empirical coverage, which can translate to improved downstream policy selection.
\end{abstract}

% we construct a principled Bayesian estimator for policy information gain that leverages a recent class of methods known as Predictive Bayes. By explicitly

\section{Introduction}
In many critical decision-making domains, acquiring complete information is often constrained by monetary cost, operational delays, or physical risks~\citep{trapeznikov12multi,janisch2020classification,erion2022cost}. This challenge is particularly prominent in clinical risk prediction~\citep{erion2022cost}. Active Feature Acquisition (AFA) frames feature acquisition as a sequential decision-making problem that adaptively selects which features to measure next with the goal of maximally reducing uncertainty about a target variable. For example, relying on principles of Bayesian Experimental Design, AFA optimizes feature selection by reducing entropy in classification tasks or variance in regression tasks \citep{ma2018eddi, covert2023learning, chattopadhyay2023variational}. Unlike static feature selection, an AFA policy dynamically adjusts its acquisition strategy based on the information gathered so far.

\begin{wrapfigure}{r}{0.49\textwidth}
\vspace{-0.2in}
  \centering
  \includegraphics[width=\linewidth]{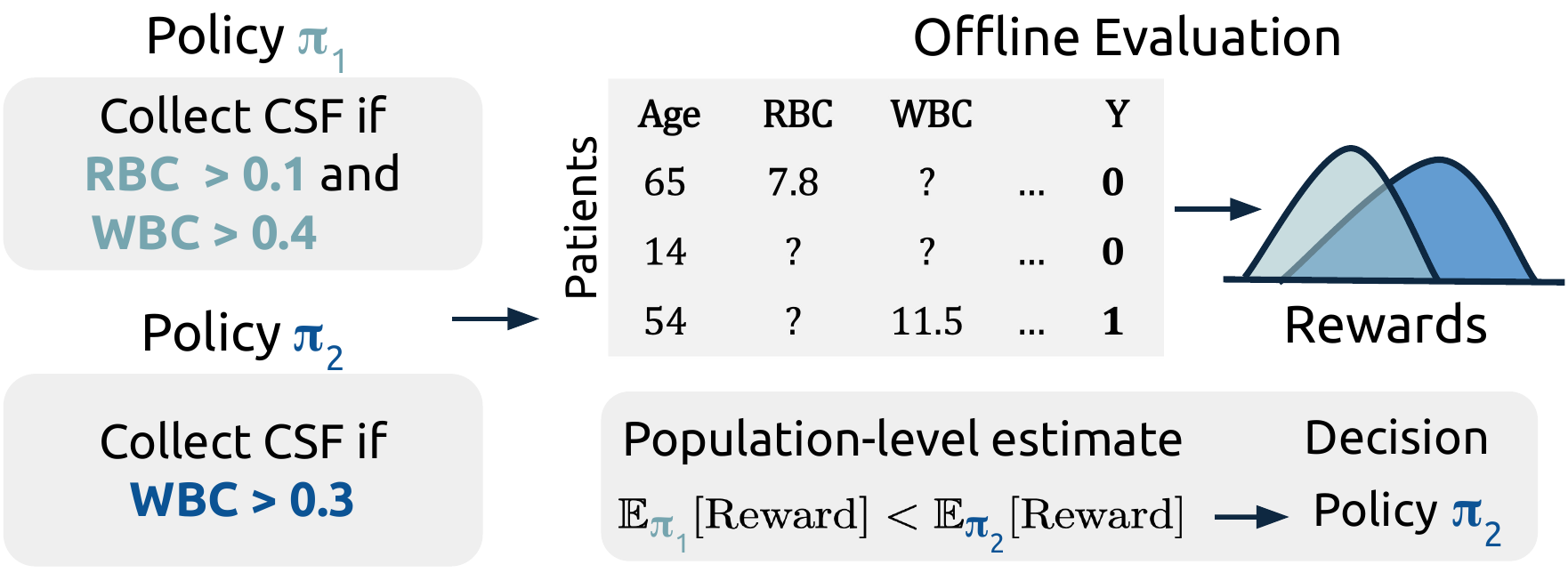}
  \vspace{-0.25in}
  \caption{Data-driven evaluation of candidate acquisition policies using historical data of covariate and outcome pairs for Alzheimer's risk prediction (CSF: cerebrospinal fluid, RBC: red blood cell count, WBC: white blood cell count).}
  \label{fig:figure1}
  \vspace{-0.1in}
\end{wrapfigure}

\emph{AFA off-policy evaluation} assesses candidate active feature acquisition policies using \emph{historical data}, collected from a different acquisition policy \citep{von2025evaluation}. Policy evaluation is a prerequisite for learning or optimizing acquisition policies. Given a retrospectively collected finite observational dataset (offline data) of feature-label pairs generated by a \emph{behavior} feature acquisition policy, the goal is to estimate the expected return (otherwise known as the \emph{policy value}) of fixed candidate \emph{target} feature acquisition policies (\textcolor{policy1}{$\pi_1$} and \textcolor{policy2}{$\pi_2$} in \Cref{fig:figure1}). 

To estimate these values, our primary contribution is a novel Bayesian formalism that enables the direct application of tabular foundation models, such as TabPFN \citep{hollmann2022tabpfn} and TabICL \citep{qu2025tabicl}. These models have demonstrated strong empirical performance and estimate posterior predictive distributions (PPDs) through a single forward pass as they are meta-learned on a diverse meta-task distribution, removing the need for explicit prior specification or extensive parameter tuning. 

We argue that robust AFA policy evaluation using PFNs necessitates decomposing the \emph{aleatoric} and \emph{epistemic} components of uncertainty to obtain a debiased estimate of the acquisition policy's value \citep{depeweg2018decomposition}. Relying on the total predictive uncertainty from a PPD estimate to evaluate feature acquisitions conflates the sources of uncertainty that can be reduced through acquisitions \citep{kobayashi2025learning}. We prove that under standard information gain objectives, this conflation underestimates feature acquisition reward by inflating uncertainty in regions of low data support. We show that, relative to the aleatoric objective, the error of the
total-uncertainty objective is exactly the epistemic uncertainty at the final state, which penalizes policies that acquire sparsely observed
features. In contrast, we recover the bias-corrected estimates of information from feature acquisitions by projecting the potential reduction in aleatoric uncertainty, leading to the Bayes-optimal selection criterion.

For example, an observational study of Alzheimer's disease risk may collect measurements $X$ and a binary outcome $Y$ under a standard-of-care protocol that relies mainly on cognitive tests and blood biomarkers, acquiring invasive cerebrospinal fluid (CSF) samples predominantly in a high-risk subgroup (see \Cref{fig:figure1}). We refer to this (potentially unknown) mechanism as the behavior policy, which introduces \emph{retrospective missingness} into historical observations. A clinical researcher may want to evaluate a candidate feature acquisition criterion (i.e., the target policy) hypothesized
to predict disease while reducing the number of patients for whom invasive CSF is required. Under the total-uncertainty objective, however, acquiring CSF may appear less informative than it is because sparse coverage in the historical data inflates the epistemic uncertainty at the resulting states.

Building on this insight, we develop a Bayesian estimation procedure to recover the posterior distribution of the expected information gain for target acquisition policies, leveraging positivity and missing-at-random (MAR) assumptions to ensure causal identification of the policy value \citep{von2025evaluation}. We instantiate this estimator using Bayesian predictive inference (BPI), which directly recovers the posterior through a sequence of one-step predictive distributions \citep{fongmartingale, fortini2025exchangeability}. This allows tabular foundation models to perform principled uncertainty decomposition, allowing us to directly estimate the aleatoric objective (posterior mean) of our AFA policy value \citep{nagler2025uncertainty, melnychuk2026frequentist}.
% \begin{figure}
%     \centering
%     \includegraphics[width=0.99\linewidth]{figures/Fig_1.pdf}
%     \caption{\emph{Left.} We consider the problem of data-driven evaluation of candidate acquisition policies using historical data of covariate and outcome pairs (CSF: cerebrospinal fluid, RBC: red blood cell count, WBC: white blood cell count). \emph{Right.} We show that the conflation of uncertainty leads to underestimation of information gain in regions of low target feature coverage. Our approach circumvents this issue by estimating the expected entropy reward over the posterior.}
%     \label{fig:figure1}
% \end{figure}

We organize the paper as follows:
\begin{itemize}[leftmargin=*]
    \item (\textbf{\Cref{sec:estimation}}) We demonstrate that using total predictive uncertainty as a surrogate reward introduces \emph{epistemic bias}. We define this bias as the systematic overestimation of uncertainty---and the consequent underestimation of information gain---for features with low support, caused by retrospective missingness in finite offline data.
    \item (\textbf{\Cref{sec:method}}) We construct a Bayesian estimator for the target policy's information gain and instantiate it via Bayesian Predictive Inference (BPI). This bypasses traditional prior specification, seamlessly integrates tabular foundation models, and mitigates epistemic bias while yielding rigorous credible intervals for policy evaluation.
    \item (\textbf{\Cref{sec:experiments}}) We empirically validate our approach across synthetic environments and real-world datasets. We show that BPI estimators leveraging specialized PFNs and pretrained foundation models (TabICL) successfully circumvent epistemic bias and provide reliable uncertainty bounds.
\end{itemize}

\section{AFA Off-Policy Evaluation}
\label{sec:setup}
% We begin in \Cref{sec:setup} by introducing the notation and problem setup for AFA policy evaluation. 
In this section, we define the target causal functional using expected entropy reduction over the feature acquisition trajectory distribution as the reward, and establish the necessary conditions for identifying this functional from observational data. 
% \textbf{Notation.} 

We denote $\mathbb{P} \in \mathcal{P}$ as the underlying true joint distribution (probability measure) over the environment, and $p$ as the corresponding density. Upper case denotes random variables, and lower case denotes instantiations. In prediction problems, one considers an environment instance $(X^*, Y, R) \sim \mathbb{P}$, where $X^{*}$ is the set of realized (observed) variables corresponding to the underlying true values $X^{(1)}$. The proxy variables $X^{*}$ are deterministically generated by $X^{(1)}$ and $R$:
$$X_j^* = \begin{cases} X^{(1)}_i & \text{if } R_j = 1 \\ \text{``?"} & \text{if } R_j = 0 \end{cases}$$
where $R \in \{0, 1\}^d$ is a set of binary missingness indicators. $Y \in \{0, 1\}$ is the target variable. While we focus on classification in the main text for clarity, our framework extends naturally to regression problems, with details deferred to \Cref{app:regression}.

When evaluating candidate acquisition policies, we assume access only to observational data. We consider paired observational and target distributions $(\Pobs, \Ptar) \in \mathcal{P}_{\mathrm{obs}} \times \mathcal{P}_{\mathrm{tar}}$. The observational distribution $\Pobs(X^*, Y, R)$ is defined as the joint distribution over the observed values, target, and missingness indicators. The target distribution $\Ptar(X^{(1)}, Y) = \Ptar(X, Y, R \mid \operatorname{do}(R=\mathbf{1}))$ corresponds to the counterfactual scenario where all features are retrospectively acquired. Although $\Ptar(X^{(1)}, Y)$ is the ideal target distribution over which the acquisition policy should be evaluated, we only have access to samples from $\Pobs$, leading to an offline policy evaluation problem.

Let the population-level functional over the target distribution be $G(\Ptar)$, where $G : \mathcal P \rightarrow \mathbb R$. This functional is identifiable if there exists a measurable function $\bar G$ such that  $G(\Ptar) = \bar G(\mathbb \Pobs)$. In other words, it can be estimated using samples from the observational distribution. To express the desired functional, we define additional random variables for the trajectory of the acquisition policy. Let $\underline{X}_t = \{X_0, ..., X_t\}$ be the set of features acquired until time $t$, and $\underline A_t = \{A_0, ..., A_{t}\}$ be the set of acquisition actions taken until $t$. We denote $S_t = (\underline X_t, \underline A_{t-1})$ as the state at $t$ for $t=1,...,T$ and $S_0 = (X_0)$ as the baseline features. We denote a trajectory of states and actions as $\tau = \{S_0, A_0, S_1,...A_{T-1},S_T\}$.

\paragraph{The Target Functional.} Standard AFA formulation considers prespecified deterministic costs $c(a_t)$ for acquiring a feature $X_{a_t}$ by acquisition action $a_t$. We consider the pointwise entropy reduction (or information gain) of $Y$ from initial state $s_0$ to terminal state $s_T$: 
\begin{align*}
H(Y \mid S_0=s_0, \mathbb{P}_{\mathrm{tar}}) - H(Y \mid S_T=s_T, \mathbb{P}_{\mathrm{tar}}).
\end{align*}
We define a target acquisition policy $\pi_\alpha : \mathcal{S} \rightarrow \Delta(\mathcal{A})$ where $\Delta(\mathcal{A})$ denotes the probability simplex over the acquisition action space $\mathcal{A}$, i.e. the unobserved features.

During policy evaluation, we seek to estimate the expected return of the target policy, which we define as the pointwise entropy marginalized over the population's trajectory distribution $S_T^{(\pi_\alpha)}$. Since the initial-state entropy is a constant across policies, the final target objective simplifies to the expected final-state entropy minus the expected acquisition costs under $\pi_\alpha$:
\begin{align}
\label{eq:objective}
G(\mathbb{P}_{\mathrm{tar}}) = 
\underbrace{- H(Y \mid S_T^{(\pi_\alpha)}, \mathbb{P}_{\mathrm{tar}}) \vphantom{\mathbb{E}_{\tau \sim \mathbb{P}_{\pi_\alpha}}\left[\sum_{t=0}^T c(A_t)\right]}}_{\text{Expected Final-State Entropy}} - 
\underbrace{\mathbb{E}_{\tau \sim \mathbb{P}_{\pi_\alpha}}\left[\sum_{t=0}^{T-1} c(A_t)\right]}_{\text{Deterministic Costs}}
\end{align}
\textbf{Identification.} Because observational data are
% we only have access to samples
generated by an unknown behavior policy $\pi_\beta~=~\Pobs (R \mid X^{(1)})$ that dictates the availability of features, we must establish conditions under which we can express $G(\Ptar) = \bar G(\Pobs)$. We rely on two standard assumptions about $\Pobs$ established in prior work \citep{von2023evaluationstatic}:
\begin{assumption}[Positivity]
\label{assumption:positivity}
    If $\mathbb{P}_{\pi_\alpha}(\tau) > 0$, then $\pi_\beta(R \geq R_T \mid X_o) > 0$, where $R_T \in \{0,1\}^d$ denotes the indicator vector for acquired features $\underline A_T$ in the trajectory $\tau$.
\end{assumption}
\vspace{-5mm}
\begin{assumption}[Missing at Random]
\label{assumption:mar}
$X^{(1)} \ci R \mid X_o$, where $X_o$ denotes the subset of features always observed, i.e. $X^{(1)} = [X_o, X_u]$
\end{assumption}
The behavior policy $\pi_\beta$ that generated $\Pobs$ needs to induce sufficient coverage of all states visited by $\pi_\alpha$. Additionally, the missingness mechanism of features depends only on the set of features that are always observed in $\Pobs$. 
Under these assumptions, we can construct a functional $\bar G(\Pobs)$ to express the cumulative reward of the target policy. We defer the identification results and the derivation of the functional to \Cref{app:identification_proof}.

\section{Challenge of Epistemic Bias in Entropy Estimation}
\label{sec:estimation}
We now focus on the entropy component of \Cref{eq:objective}, where we wish to apply a Bayesian PPD estimator such as PFNs using an offline finite dataset $z_{1:n} \sim \Pobs$. This requires inferring the targets $\theta_{s_T}$, which parameterize the underlying true Bernoulli distributions $\tilde {\mathbb P} \coloneqq \mathbb{P}_{\theta_{s_T}}(Y \mid s_T)$ respectively.

Following \citet{jayasekera2026variational}, we decompose the total predictive entropy after observing $z_{1:n}$:
\begin{align}   
H(Y \mid s_T, z_{1:n}) &= \underbrace{\mathbb{E}\left[H(Y \mid s_T, \tilde{\mathbb P}) \mid z_{1:n} \right]}_{\text{Aleatoric uncertainty}~U_a(s_T)} + \underbrace{I(Y; \tilde{\mathbb P} \mid s_T, z_{1:n}) \vphantom{\mathbb{E}\left[H(Y \mid s_T, \tilde{\mathbb P}) \mid z_{1:n} \right]}}_{\text{Epistemic uncertainty}~U_e(s_T)} 
% = U_a(s_T) + U_e(s_T) 
\label{eq:aleatoric_decomp}
\end{align}
where $U_e(s_T) = I(Y; \tilde{\mathbb P} \mid s_T, z_{1:n}) = \mathbb{E}\left[ D_{\mathrm{KL}} \Big( \mathbb{P}(Y \mid s_T, \tilde{\mathbb P}) \parallel \mathbb{P}(Y \mid s_T, z_{1:n}) \Big) \mid z_{1:n} \right]$ is the uncertainty about the true parameter that can be resolved as we collect more samples. $U_a(s_T)$ is the irreducible uncertainty in $Y$ given $s_T$. This stems either from intrinsic noise in the target variable or from unobserved features missing from the current state $s_T$ that explain the variation in $Y$.

As $n \rightarrow \infty$, the posterior distribution $ \Pi(\tilde{\mathbb P}\mid z_{1:n})$ over the conditional measure converges weakly to a Dirac measure at the true parameter $\delta(\theta_{S_T})$ \citep{ghosal2017fundamentals, fongmartingale}. In this infinite-data limit, epistemic uncertainty vanishes ($U_e \rightarrow 0$). However, under finite data $z_{1:n}$, using the total predictive entropy systematically overestimates the conditional entropy by an amount exactly equal to the epistemic uncertainty $I(Y ; \tilde{\mathbb P} \mid s_T, z_{1:n}) \geq 0$ in expectation under the posterior.

% We consider the information gain, defined as the expected entropy reduction from $s_0$ to $s_T$. Using our previous decomposition in \Cref{eq:aleatoric_decomp}, we can express the systematic bias as:
% \begin{equation}  
% \label{eq:decomp}
% H(Y \mid s_0, z_{1:n}) - H(Y \mid s_T, z_{1:n}) = \underbrace{\Big( U_a(s_0) - U_a(s_T) \Big)}_{\text{Posterior Expected Information Gain}} + \underbrace{\Big( U_e(s_0) - U_e(s_T) \Big)}_{\text{Epistemic Bias}}
% \end{equation}

In observational datasets subject to retrospective missingness, some acquisitions may have lower support in offline data (e.g., invasive CSF, PET scans). Consequently, $U_e(s_T)$ can be large. Relying on total predictive uncertainty $H(Y \mid s_T, z_{1:n})$ in \Cref{eq:aleatoric_decomp} can thus systematically \emph{overestimate} uncertainty in regions with poor data coverage. This leads to suboptimal policy selection if the optimal features are not sufficiently represented in the offline dataset. We characterize the regret in \Cref{app:regret}. 

\section{Bayesian Predictive Inference for AFA Value Estimation}
\label{sec:method}
In this section, we present our framework for estimating the AFA policy value. By explicitly estimating the aleatoric component of \Cref{eq:aleatoric_decomp} under the target policy, our approach guarantees a bias-corrected reward estimate, unconfounded by $U_e$. \Cref{sec:pred_bayes} introduces Bayesian predictive inference (BPI). This framework derives posterior distributions for quantities of interest directly from a posterior predictive distribution (PPD). %, which can be readily obtained from off-the-shelf foundation models. 
Since modern foundation models are typically trained with an autoregressive loss, using data samples as context trained with a next-token prediction autoregressive loss effectively yields a PPD estimator. For instance, TabPFN~\citep{muller2021transformers} is an off-the-shelf foundation model that can be used as a PPD estimator combined with in-context learning to adapt to new tasks. 
\Cref{sec:value} details how we compute the posterior over the finite sample estimate of the target AFA functional, $\bar G(\Pobs)$, to estimate the policy value.

\subsection{Bayesian Predictive Inference for Estimating $U_a$} 
\label{sec:pred_bayes}
Standard Bayesian estimation requires explicit specification of a prior-likelihood pair, which is often difficult to design in practice. Instead, we investigate BPI, which performs Bayesian estimation using a sequence of one-step-ahead predictive distributions (such as PPDs) defined directly on observable quantities. Under this framework, statistical uncertainty arises strictly from finite sample sizes. With infinite observations, the quantity of interest could be identified precisely. Given $n$ samples, however, the samples unobserved so far ($n+1 \rightarrow \infty$) are the sole source of $U_e$.

Let $Z \in \mathcal{Z}$ be a generic random variable with distribution $\mathbb{P}(Z)$, and let $F$ denote the CDF. Assume we have access to a sequence of one-step-ahead predictive distributions $P_i~=~P(Z_{i+1} \in \cdot | Z_{1:i})$ for $i\geq 1$ and $P_0 = P(Z_1\in \cdot)$. If the sequence of $(Z_n)_{n \geq 0}$ is infinitely exchangeable, \emph{de Finetti's representation theorem} guarantees the existence of a limiting random probability measure $F_\infty$ on $\mathcal Z$, and $(P_i)_{i>0}$ converges almost surely to $F_\infty$ \citep{de1937prevision, fortini2025exchangeability}. Intuitively, $F_\infty$ acts as the underlying data-generating measure. Given finite data $z_{1:n}$, the posterior distribution is given by $\Pi(F_\infty \mid z_{1:n})$, where the variation in the limit $F_\infty$ is due to $U_e$ given finite samples. We outline two approaches for obtaining the limiting measure of $Y \mid X$ under this framework, which we describe in more detail in \Cref{app:posterior_algorithms}. 

\paragraph{Martingale Posterior \citep{fongmartingale, ng2025tabmgp}.} The existence of $F_\infty$ can be achieved through a weaker sufficient condition known as the martingale property (see \Cref{eq:martingale}). This guarantees that the sequence of predictive distributions $(P_i)_{i \geq 0}$ is a bounded martingale and converges as $i \rightarrow \infty$ to a random limit $F_{\infty}$. This motivates instantiating a sequence of predictives and simulating pseudo-observations autoregressively, known as \emph{predictive resampling}. Predictive resampling (i.e. repeatedly drawing the next observation conditional on observations so far) yields an approximate Monte Carlo sample of $F_\infty | z_{1:n}$. Collecting these limits yields a random probability measure over $F_\infty | z_{1:n}$ which can be viewed as the posterior distribution. Prior works have explored using PFNs \citep{ng2025tabmgp} and copulas \citep{fongmartingale} and their combinations \citep{melnychuk2026frequentist, nagler2025uncertainty} as the sequence of predictives, with the latter satisfying the martingale property by construction.

% \begin{figure}
%     \centering
%     \vspace{-0.1in}
%     \includegraphics[width=\linewidth]{figures/Methods_figure2.pdf}
%     \caption{We show that the conflation of uncertainty leads to underestimation of  information gain in regions of low target feature coverage. Our approach circumvents this issue by estimating the expected entropy reward over the posterior.}
%     \label{fig:methods}
%     \vspace{-0.1in}
% \end{figure}

\begin{figure}
    \centering
    \vspace{-0.1in}
    \includegraphics[width=\linewidth]{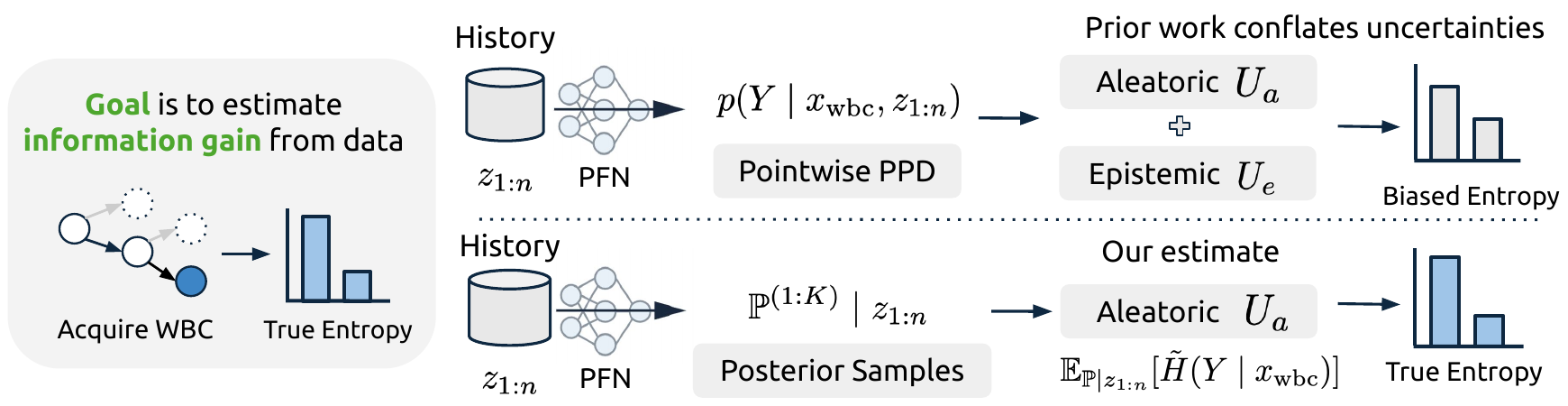}
    \vspace{-0.1in}
\caption{\textbf{Uncertainty Decomposition for Value Estimation of AFA Policies.} \textbf{Top:} Using pointwise PPD ($p(Y \mid x_{\text{wbc}}, z_{1:n})$) conflates aleatoric ($U_a$) and epistemic ($U_e$) uncertainties and leads to underestimation of information gain. \textbf{Bottom:} Our approach resolves this issue by drawing posterior samples ($\mathbb{P}^{(1:K)} \mid z_{1:n}$) to  estimate the expected entropy over the posterior ($\mathbb{E}_{\mathbb{P} \mid z_{1:n}}[\tilde{H}(Y \mid x_{\text{wbc}})]$), isolate aleatoric uncertainty $U_a$.}
    \label{fig:methods}
    \vspace{-0.1in}
\end{figure}

\paragraph{Predictive Central Limit Theorem (CLT) \citep{fortini2026principled}.} The predictive CLT is an alternative approach that bypasses predictive resampling and forms an estimate by viewing the probability measure over $F_\infty | z_{1:n}$ as asymptotically Gaussian under certain assumptions (\Cref{app:clt}). Let $P_{k, i} = P(Y=B_i \mid X_{k+1}=x_i, Z_{1:k})$ denote the predictive probability for a given measurable $B_i \subseteq \mathcal Y$ and covariates $x_i$. We then define $\mathbf P_k \in [0,1]^{m}$ as the vector of these predictive probabilities over $m$ distinct covariate-event pairs.
\begin{align}
\label{eq:clt_estimator}
    P_\infty(\cdot) \mid z_{1:n} \sim \mathcal N_m(\mathbf{P}_n, \frac{V_n}{n}) , \quad V_n=\frac{1}{n}\sum_{k=1}^n k^{2}\Delta_k\Delta_k^T.
\end{align}
The covariance estimate $V_n$ is formed by the volatility of the predictive sequence $\Delta_k = \mathbf{P}_{k} - \mathbf{P}_{k-1}$ from the observations so far ($k=1$ to $n$), instead of simulating future observations. Intuitively, the goal is to estimate the remaining random distance from $F_n$ to the limiting measure $F_\infty$, given by the sum of the unobserved future step-wise volatility $F_\infty - F_n = \sum_{k=n+1}^\infty \Delta_k$. \Cref{eq:clt_estimator} is a practical estimator for the variance $\text{Var}(F_\infty - F_n \mid Z_{1:n})$ proposed in \cite{fortini2026principled}. 

\paragraph{Entropy Decomposition.} Once we estimate the posterior $\Pi(F_\infty \mid z_{1:n})$ using the methods described above, we can evaluate the aleatoric component of the entropy in \cref{eq:aleatoric_decomp}. We compute this as the expected entropy of the limiting measure $F_\infty$: $U_a(s_T) = \mathbb{E}\left[H(Y \mid s_T, F_\infty) \mid z_{1:n} \right]$. The following section builds upon this instance-level estimate to derive the expected population-level $U_a$ for a target acquisition policy.

\subsection{Bayesian Value Estimation for Feature Acquisition Policies} 
\label{sec:value}
We now outline our procedure for deriving posteriors over the AFA policy value functional $\bar G(\Pobs) \in \mathcal G$. We first review traditional estimation methods for obtaining a point estimate of $\bar G(\Pobs)$, which requires estimating $\{\mu, \pi_\beta\}$ where $\mu = \mathbb P(Y \mid S_T)$ and $\pi_\beta = \mathbb P(R \mid X_o)$. We then extend these estimators to their Bayesian counterparts by defining independent posteriors over $\{\mu, \pi_\beta\}$. Using the methods outlined in \Cref{sec:pred_bayes}, our approach generates functional draws from each posterior distribution. We then construct the posterior over the population-level policy value via the pushforward: 
\begin{align*}
\Pi(G(F_\infty) \in A \mid z_{1:n}) = \int \mathbf{1}(G(F_\infty) \in A) \, d\Pi(F_\infty \mid z_{1:n}).
\end{align*}
\textbf{Frequentist Estimators.} 
% We begin by outlining the traditional method for obtaining a point estimate of . 
We consider the Inverse Propensity Weighting (IPW) estimator to obtain a point estimate of $\bar G(\Pobs)$ \citep{von2025evaluation}. We define the following trajectory-level importance weight:
% Instead of estimating the full joint distribution $\Pobs$, the IPW estimator only requires the following set of nuisances $\{\mu, \pi_\beta\}$ where $\mu = \mathbb P(Y \mid S_T)$ and $\pi_\beta = \mathbb P(R \mid X_o)$.
% We define the following trajectory-level importance weight:
\begin{align*}
    w_T(\tau) \coloneqq \frac{\mathbb I[R \geq R^T]\prod_{t'=0}^T\pi_\alpha(A_{t'} \mid S_{t'})}{\pi_\beta(R \geq R^T\mid X_o)\prod_{t'=0}^T\pi_{\mathrm{sim}}(A_{t'} \mid S_{t'}, R)},
\end{align*}
where $\pi_{\mathrm{sim}}$ acts a a simulator that induces a distribution over trajectories, $d_{S_T} \coloneqq \mathbb P(\underline X_T, \underline A_{T-1} \mid R, X^*)$. We defer the specific design choices for this simulator to \Cref{app:identification_proof} with additional considerations to avoid explicit covariate density estimation (e.g., via imputation). Using this weight, we can estimate the target policy's expected entropy as:
\begin{align}
\label{eq:estimator_ipw}
    \bar G(\Pobs) = \mathbb{E}\left[w_T(\tau)H(Y \mid S_T) \right],
\end{align}
where the expectation is taken over the observational data augmented with the simulated acquisition trajectory, distributed according to $\mathbb{P}(\underline X_T, \underline{A}_{T-1}, R, X^*)$.

\paragraph{Obtaining Functional Posteriors.} 
By placing independent priors on the unknown components $\mu$, $\pi_\beta$, and $d_{S_T}$, the resulting posterior factorizes, allowing us to sample each component conditionally independently given the data $z_{1:n}$ \citep{ray2020semiparametric}. We sample $\pi_\beta$ and $\mu$ from the functional posterior using the procedures outlined in \Cref{sec:pred_bayes}. For $d_{S_T}$, we avoid parametric density estimation by employing a Bayesian Bootstrap procedure (\Cref{app:bayesian_bootstrap}) to obtain a nonparametric posterior for this marginal distribution \citep{rubin1981bayesian}. Using Monte Carlo draws from these functional posteriors, we estimate the aleatoric entropy under the target policy in ~\Cref{eq:estimator_ipw} by computing the posterior mean of the functional:
\begin{align*}
    \tilde G(\Pobs) &= BB_n\{\tilde w_T(\tau)\tilde H(Y \mid S_T) \};\quad \tilde \pi_\beta, \tilde \mu \sim \Pi(\pi_\beta, \mu \mid z_{1:n}) \\
    \hat{U}_a(S_T) &= \mathbb{E}_{F_\infty \mid z_{1:n}}\Big[\tilde G(\Pobs)\Big]. 
\end{align*}
where $BB_n$ denotes the mean wrt. the Bayesian Bootstrap process $\sum_{i=1}^n W_if(z_i)$ with $(W_1,...,W_n)$ sampled from a uniform Dirichlet process.

% \textbf{Frequentist Consistency.} While BPI provides functional posterior samples, we are additionally interested in whether this Bayesian estimation procedure satisfies frequentist consistency. This property is desired, as it ensures asymptotically that (1) the result is indifferent to prior specification and (2) agreement between Bayesian and frequentist estimators. Therefore, we apply a one-step posterior correction (OSPC) to our estimation procedure for the posterior $\mathbb{P}(\psi^{OSPC}(\eta) \mid \mathcal D)$. 

\section{Related Work}
Active Feature Acquisition (AFA) formulates feature selection as a sequential decision-making process, where the goal is to learn policies that dynamically select which features to acquire next to optimally predict a target variable. Previous studies have predominantly focused on \emph{optimization} of task-specific acquisition for a static offline dataset \citep{shim2018joint, kachuee2019opportunistic, janisch2019classification}, and we focus on approaches that define the learning objective based on greedy or multi-step information gain \citep{covert2023learning, gadgil2023estimating, ghosh2023difa, kobayashi2025learning}. To estimate the optimal acquisition policy, existing methods typically rely on either generative sampling-based techniques \citep{ma2018eddi} or policy networks that directly output the optimal next action \citep{covert2023learning, guney2025active, aronsson2026non}. In contrast, our work focuses strictly on the \emph{evaluation} of pre-defined acquisition policies under offline data with inherent missingness \citep{von2025evaluation}. Rather than proposing a new optimization method, our goal is to formulate and validate estimators for the policy value driven by tabular foundation models (PFNs). This integration provides a highly practical and scalable inference mechanism for offline data, yielding strong empirical performance without the overhead of tuning bespoke models.

% \textbf{Off-Policy Evaluation and Selection under Uncertainty.} Off-policy evaluation focuses on estimating a target policy's value using offline data collected from a behavior policy, with the goal of selecting a policy for deployment. Various estimation strategies have been proposed. There are frequentist \citep{thomas2015high, kostrikov2020statistical} and Bayesian \citep{yang2022offline} formulations for providing confidence estimates in the target policy performance, as opposed to point estimates. 

\section{Experiments}
\label{sec:experiments}
\textbf{Evaluation Metrics.} We evaluate our approach on two primary dimensions: the quality of the uncertainty estimates for the policy value, and the accuracy of downstream acquisition policy selection. We define a \emph{ground-truth reference value} either analytically using closed-form expressions in simulations, or empirically using a large oracle dataset without missingness.

\begin{itemize}[leftmargin=*, noitemsep, topsep=0pt, parsep=0pt]
    \item \textbf{Bias of point estimate:} The error relative to the reference value determines whether our aleatoric entropy estimator successfully mitigates epistemic bias.
    \item \textbf{Coverage assessment:} A standard diagnostic for uncertainty quantification is frequentist coverage \citep{ng2025tabmgp}, which verifies whether the epistemic uncertainty captured by the posterior reliably bounds the reference estimand. The resulting empirical coverage rate of the credible intervals should be close to $1-\alpha$ where we select $\alpha=0.05$. We also report the interval width.
    \item \textbf{Ranking and regret of candidate acquisitions:} We measure the policy regret and the relative accuracy of action selection compared to the reference estimand, reporting pairwise accuracy (comparing the relative ordering accuracy of all candidate pair combinations).
\end{itemize}

We defer additional evaluation of epistemic uncertainty estimates for the conditional $Y \mid S_T$ using various known diagnostics such as posterior contraction \citep{ng2025tabmgp} to \Cref{app:diagnostics}. 

\paragraph{Baselines.} We evaluate six estimators across two distinct categories:
\begin{itemize}[leftmargin=*, noitemsep, topsep=0pt, parsep=0pt]
\item \textbf{Total uncertainty estimators:} These variants yield a biased point estimate for the true entropy. (1) \textbf{Total (plug-in)} uses a single PFN pass to obtain a PPD estimate $P_n$ after $n$ observations and plugs it directly into the entropy estimation.
(2)~\textbf{Total (bootstrap)} resamples the context $z_{1:n}$ to refit the PFN $K$ times and averages predictions before calculating entropy.
\item \textbf{Bias-corrected aleatoric estimators:} These variants (including both non-BPI and BPI approaches) mitigate epistemic bias and generate credible intervals. 
(3) \textbf{Bootstrap} performs the same $K$ refits but computes the entropy \emph{within} each sample before averaging. 
(4) \textbf{Marginal} is an ablation using the BPI posterior but restricting it to the marginal rather than the joint distribution. 
(5) \textbf{Copula} implements the BPI posterior using the copula martingale posterior. 
(6) \textbf{CLT} implements the BPI posterior using the predictive CLT.
\end{itemize}
% We restrict the comparison to these approaches that directly use an off-the-shelf foundation model, as deep ensembles, MC dropout, and Bayesian neural networks (BNN) require model specification and retraining. 

\subsection{Synthetic Datasets}
\textbf{Data Generating Process.} We consider settings where the true conditional distribution $\mathbb P(Y \mid S_T)$ is exactly computable for any partially observed input $S_T$. This provides the ground-truth conditional entropy needed to directly assess the quality of policy evaluation. We evaluate three classes of DGPs: (1) \textbf{Linear}: Each feature is  given by $X_j \mid Y \sim \mathcal N(\mu_{j, y}, 1)$. (2) \textbf{GMM}: Each class is a mixture of isotropic Gaussians, $X \mid Y \sim \sum_k w_{k,y} \mathcal N(\mu_{k, y}, \sigma I)$, leading to nonlinear decision boundaries. (3) \textbf{BNN}: A random one-hidden-layer MLP, where $X \sim \mathcal N(0, I)$ and $Y \mid X \sim \text{Bernoulli}(\sigma(f(X)))$. The true signal strength is varied across features. Full experiment details are in \Cref{app:synthetic}. 

For evaluation, we generate 200 distinct DGPs per class by randomly sampling their underlying parameters (e.g., target prevalence, Gaussian moments, or BNN weights). For each DGP, we sample a training set size $n_{\text{train}}$ uniformly at random, and generate 5 replications of varying total missingness rates from $\mathbb P(X^*, R, Y)$. To introduce missingness, we define a random linear Missing At Random (MAR) mechanism $\mathbb P(R \mid X_o)$ where the rate varies across features, leading to imbalanced coverage. 

\paragraph{Model.} For each class of DGPs, we train a PFN to approximate the PPD of partially observed query states $P(Y \mid S_T, Z_{1:n})$ on datasets sampled from the possible range of DGPs. For this evaluation, we train PFNs from scratch instead of relying on off-the-shelf pretrained models to reduce the possible confounding of PPD estimation error. Our goal is to verify whether a PFN that is trained close to the idealized Bayesian limit can recover accurate value estimates and select optimal acquisition actions.

\begin{figure}[t]
    \centering
    \includegraphics[width=\linewidth]{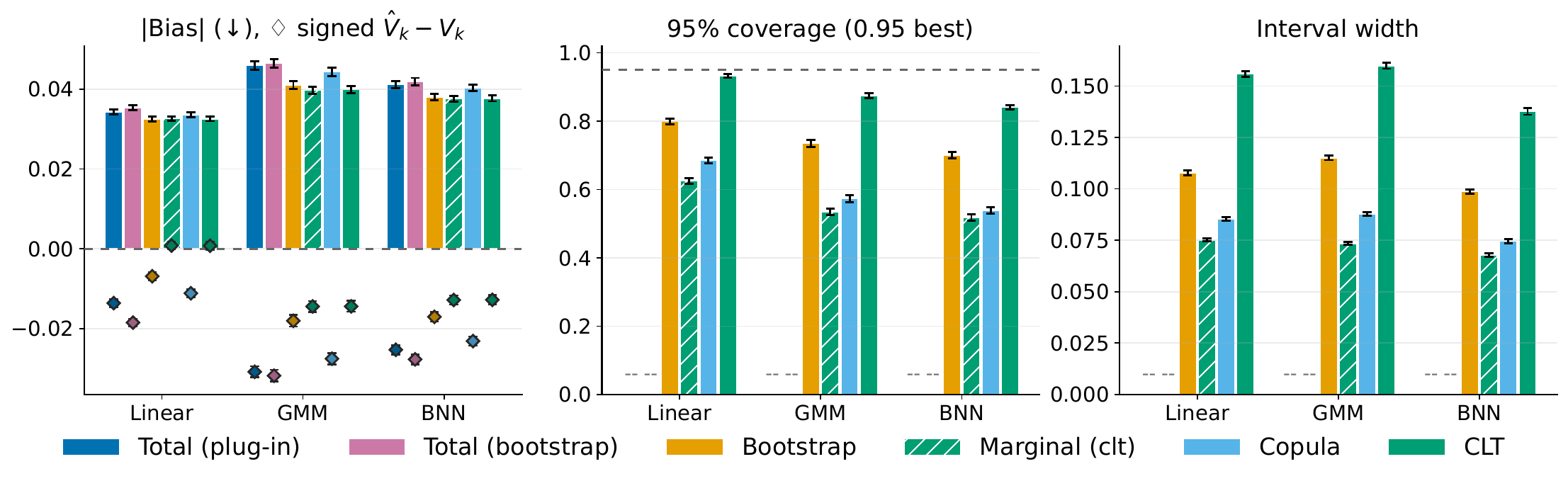}
    \vspace{-0.25in}
\caption{\textbf{Value bias and coverage for synthetic DGPs.} Value estimates averaged over 1000 datasets sampled per simulated DGP class (mean $\pm$ standard error). The \textbf{Total} estimators suffer from epistemic bias leading to underestimation of the true feature acquisition value (negative signed bias), whereas the proposed BPI and Bootstrap approaches mitigate this bias.}
        \label{fig:bias}
        \vspace{-0.1in}
\end{figure}

\paragraph{Improved Value Estimation Compared to Total Predictive Uncertainty.} We characterize the bias of the point estimate of information gain for the fully-synthetic DGPs in ~\Cref{fig:bias} (left). The target policy is a one-step acquisition that selects a random subset of informative features for all samples, starting from a baseline state. The epistemic underestimation in the predictive uncertainty at $P_n$ is apparent for the \textbf{Total} estimators, where the average bias is consistently the most negative in both the plug-in and bootstrap variants. We see the intended mitigation effect using the aleatoric estimators which mitigates this negative bias and also achieves lower absolute bias, averaged across datasets.

\paragraph{Undercoverage of the Marginal and Bootstrap Estimators.} 
Comparing the \textbf{Marginal} and \textbf{CLT} estimators in \Cref{fig:bias} reveals that while they yield almost identical pointwise estimates (and thus the same bias), the joint sampling approach significantly improves coverage. This demonstrates that obtaining functional draws of $Y \mid S_T$ is essential for recovering the true uncertainty of the population-level functional $G(\Pobs)$. Compared to the \textbf{Bootstrap} baseline, \textbf{CLT} achieves superior coverage by propagating epistemic uncertainty, resulting in appropriately wider confidence intervals. 

In contrast, \textbf{Copula} exhibits undercoverage. We attribute this to the copula's tendency to underestimate uncertainty in regions with poor covariate overlap. We verify this phenomenon in Appendix~\ref{app:diagnostics}, which aligns with findings in prior work \citep{fongmartingale, ng2025tabmgp}.

\begin{figure}[t]
    \centering
    % \vspace{-0.15in}
    \includegraphics[width=\linewidth]{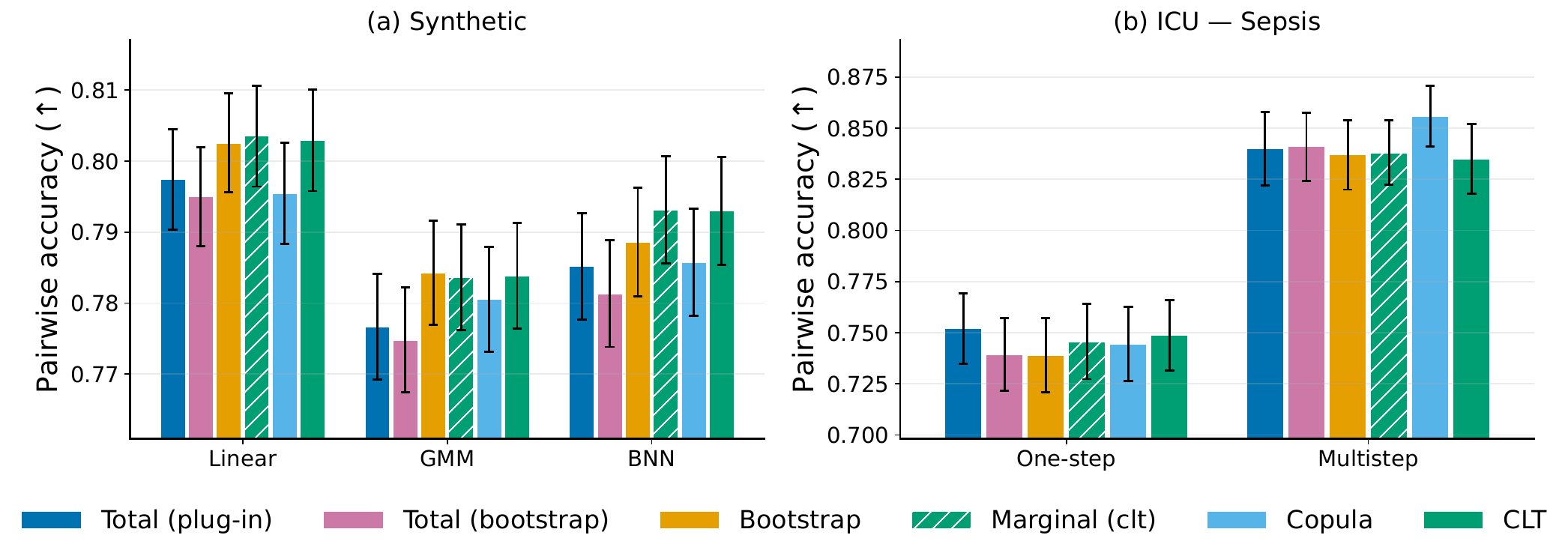}
    \vspace{-0.28in}
    \caption{\textbf{Downstream action selection performance on synthetic and real-world environments.} We report pairwise accuracy for the simulated DGPs and clinical tasks (sampled across 200 and 100 datasets per setting, respectively). Within each sampled dataset, we evaluate candidate policies, where single-step policies are single features over the whole population, and multistep policies are stochastic policies over sequential steps. Pairwise accuracy represents the proportion of times the estimator correctly ranks the superior policy when comparing all possible pairs of candidates.}
    \label{fig:ranking}
    \vspace{-0.12in}
\end{figure}

\paragraph{Improvements in downstream action selection.} \Cref{fig:ranking} (a) evaluates estimator performance based on pairwise policy ranking accuracy. Detailed dataset-level improvements and regret incurred from selecting suboptimal actions, averaged by missingness level, are provided in \Cref{tab:policy_ranking_vs_plugin_synthetic_linear,tab:policy_ranking_vs_plugin_synthetic_gmm,tab:policy_ranking_vs_plugin_synthetic_bnn}. Across all synthetic DGP families, the aleatoric estimators consistently match or outperform their biased counterparts. They yield statistically significant improvements in pairwise accuracy while maintaining stable regret, confirming the absence of catastrophic errors. Crucially, the unbiased estimators provide the largest performance gains under high retrospective missingness, a regime where coverage imbalance leads to misleading entropy estimates.

\subsection{Real-World Datasets}
% We evaluate our proposed approach on real-world datasets.
\textbf{Tabular Benchmarks.} We benchmark the estimators under a variety of covariate and label distributions using OpenML tabular datasets. We introduce synthetic MCAR missingness of varying rates to each feature, and evaluate single-step policy candidates that each acquire a random subset of features from zero. Experiment details are in \cref{app:tabular} with additional results in \cref{app:tabular_results}.

\paragraph{ICU Prediction.} We evaluate our approach on the \textbf{Sepsis} task from the Physionet challenge \citep{ reyna2019early}, which exhibits inherent feature missingness characteristic of retrospective clinical data. Given that the missingness mechanism is unknown, we fit a linear propensity model for $\pi_\beta(R \mid X_o)$ where we assume $X_o$ consists of patient demographics and vital signs. To simulate decision-making at arbitrary points in a patient's clinical trajectory, we assess policies corresponding to (i) all possible single-step acquisition actions from randomly sampled initial states and (ii) stochastic multistep policies from a baseline state. We compute a reference value using the full data, and obtain estimates using subsampled datasets at $n=\{500,1000,1500\}$. Full details regarding data processing and experimental setup are deferred to \Cref{app:ehr}. 

% \textbf{Model.}
% However, it remains unclear whether TabICL was explicitly trained to output PPDs for partially observed queries, and whether they are well-calibrated for our specific real-world tasks. Consequently, while our simulations confirm the theoretical advantages of our framework, our real-world results may be constrained by the current limitations of TabICL in producing ideal PPDs for accurate aleatoric entropy estimation.

% \subsection{Results}

\begin{figure}[!t]
    \centering
    % \vspace{-0.05in}
    \includegraphics[width=\linewidth]{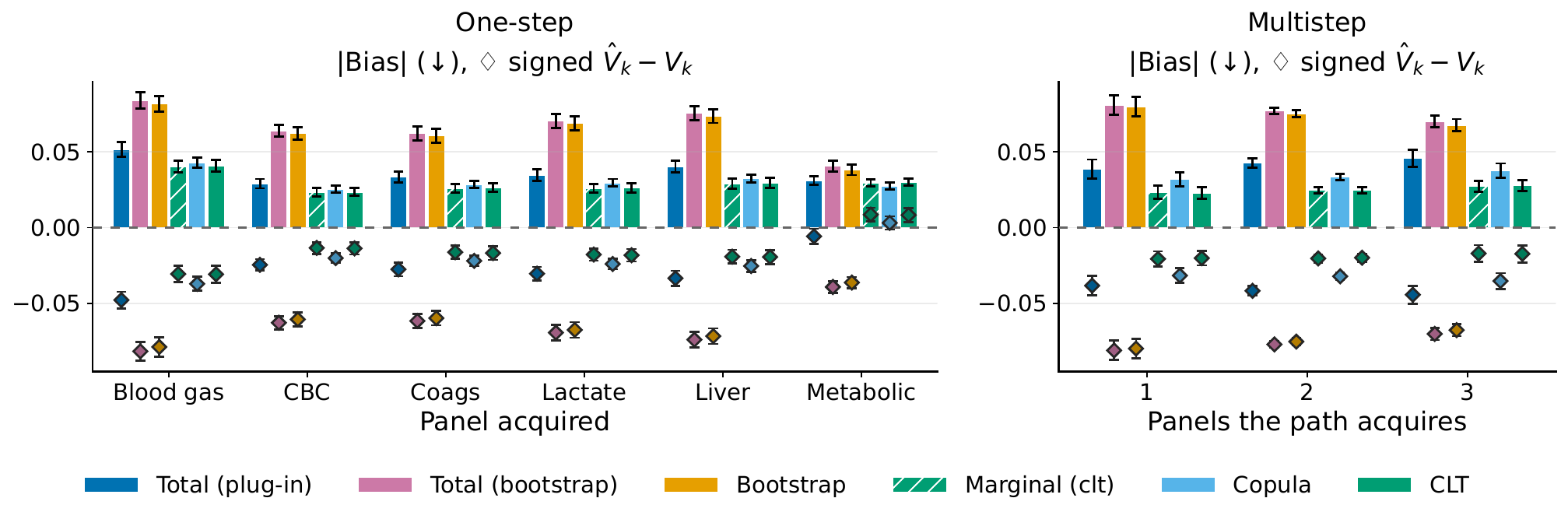}
    \vspace{-0.34in}
    \caption{\textbf{Value bias for single and multistep acquisition policies on Sepsis task.} We report the value bias across each panels for single-step acquisition, and bias across multistep sequential policies that acquire different number of panels.}
    \label{fig:icu_bias}
    \vspace{-0.15in}
\end{figure}

\paragraph{Bias Correction in Real-world Datasets.} 
We use TabICL-V2 \citep{qu2025tabicl} as the underlying off-the-shelf foundation model for PPD estimation, due to its inference speed and its transparent handling of missing features. 
\Cref{fig:icu_bias} shows analogous results for \textbf{Sepsis}, where the estimation bias is visualized across feature groups. We observe feature-specific epistemic bias in the \textbf{Total} estimators; specifically, imbalanced coverage affects entropy estimation, leading to varying levels of underestimation across feature groups. \Cref{fig:ehr_propensity} shows measurement propensity per group). In this setting, the aleatoric estimators achieve lower bias than the total predictive uncertainty baselines. 

The \textbf{CLT} estimator also yields strong coverage, with intervals only slightly wider than \textbf{Bootstrap} (\Cref{fig:icu_coverage,fig:icu_width} in the appendix). However, as shown in \Cref{fig:ranking} (b), downstream policy selection does not improve despite this debiasing. Nevertheless, we view these results as encouraging because the posterior spread of the aleatoric estimators tracks the underestimation of the plug-in value (\cref{fig:icu_spread} in the appendix). This demonstrates that they successfully capture the epistemic component of the bias. The lack of improvement in policy rankings likely occurs because the estimated bias corrections lack the precision necessary to overcome the existing value gaps between policies and alter their relative order.

% \begin{figure}
%     \centering
%     \includegraphics[width=\linewidth]{figures/EHR_figure.pdf}
%     \vspace{-0.25in}
%     \caption{Performance metrics (bias, coverage, and interval width) for value estimates averaged over 150 datasets sampled per EHR task (mean $\pm$ standard error).}
%     \label{fig:ehr_bias}
%     % \vspace{-0.2in}
% \end{figure}

\section{Conclusion} 
\paragraph{Summary.} In this work, we showed that naive Bayesian approaches using total predictive uncertainty yield biased estimates for feature acquisition. We introduced a novel Bayesian estimation framework that explicitly mitigates this bias, allowing tabular foundation models to be used for AFA policy evaluation. In synthetic settings, we demonstrate that predictive Bayesian estimators yield highly accurate value estimates and downstream improvements in policy ranking. Further, we evaluate on real-world datasets, confirming that our TabICL-based estimators mitigate epistemic bias while providing reliable uncertainty bounds.

\paragraph{Discussion and future work.} \emph{(i)} We focus on small- to medium-sized datasets where imbalanced coverage naturally yields high epistemic uncertainty, and where current PFN models can reliably produce calibrated PPD estimates; extension to larger scales is left to future work. \emph{(ii)} Because our primary goal is studying the biasing effect of epistemic uncertainty on policy value estimation, we rely on a standard inverse probability weighting (IPW) estimator. Integrating our Bayesian PFN-based strategy with advanced semiparametric estimators (e.g., the Double Reinforcement Learning estimator from \citet{von2025evaluation, melnychuk2026frequentist}) that offer stronger asymptotic guarantees, or relaxing current data-generating assumptions to handle flexible missingness mechanisms and temporality, are natural directions for future research. \emph{(iii)} While our simulations confirm the theoretical advantages of our framework, our real-world results are constrained by the accuracy of the reference values and the current limitations of TabICL in producing ideal PPDs for aleatoric entropy estimation. We expect future improvements in tabular foundation models to directly translate to reduced estimation error in our framework.

%%%%%%%%%%%%%%%%%%%%%%%%%%%%%%%%%%%%%%%%%%%%%%%%%%%%%%%%%%%%
\newpage
% \input{Statements}



\bibliography{iclr2027_conference.bib}
\bibliographystyle{iclr2027_conference}

\appendix
\newcommand{\Normal}{\mathcal{N}}
\newcommand{\Ex}{\mathbb{E}}
\newcommand{\Hent}{\mathrm{H}}
\newcommand{\Ind}{\mathbf{1}}
\newcommand{\cflag}[1]{\texttt{\small #1}}
\renewcommand\thefigure{\thesection.\arabic{figure}}    
\renewcommand\thetable{\thesection.\arabic{table}}

\section*{Appendix Contents}
\begingroup
\setlength{\parskip}{2pt}\setlength{\parindent}{0pt}
\textbf{\hyperref[app:proofs]{A\quad Theoretical Insights and Proofs}} \dotfill \pageref{app:proofs}\par
\hspace*{1.5em}A.1\quad \hyperref[app:identification_proof]{Identification Of Target Policy Value} \dotfill \pageref{app:identification_proof}\par
\hspace*{1.5em}A.2\quad \hyperref[app:regression]{Extension to Regression} \dotfill \pageref{app:regression}\par
\hspace*{1.5em}A.3\quad \hyperref[app:regret]{Regret} \dotfill \pageref{app:regret}\par
\textbf{\hyperref[app:posterior_algorithms]{B\quad Methods for Obtaining Posteriors}} \dotfill \pageref{app:posterior_algorithms}\par
\hspace*{1.5em}B.1\quad \hyperref[app:bayesian_bootstrap]{Bayesian Bootstrap} \dotfill \pageref{app:bayesian_bootstrap}\par
\hspace*{1.5em}B.2\quad \hyperref[app:martingale_copula]{Martingale Posterior via Copulas} \dotfill \pageref{app:martingale_copula}\par
\hspace*{1.5em}B.3\quad \hyperref[app:clt]{Predictive CLT} \dotfill \pageref{app:clt}\par
\textbf{\hyperref[app:experiments]{C\quad Experiments}} \dotfill \pageref{app:experiments}\par
\hspace*{1.5em}C.1\quad \hyperref[app:synthetic]{Synthetic Experiments} \dotfill \pageref{app:synthetic}\par
\hspace*{1.5em}C.2\quad \hyperref[app:tabular]{Tabular Benchmarks} \dotfill \pageref{app:tabular}\par
\hspace*{1.5em}C.3\quad \hyperref[app:ehr]{ICU lab acquisition under real-world missingness} \dotfill \pageref{app:ehr}\par
\textbf{\hyperref[app:results]{D\quad Additional results}} \dotfill \pageref{app:results}\par
\hspace*{1.5em}D.1\quad \hyperref[app:synthetic_results]{Synthetic Experiments} \dotfill \pageref{app:synthetic_results}\par
\hspace*{1.5em}D.2\quad \hyperref[app:tabular]{Tabular Benchmarks} \dotfill \pageref{app:tabular_results}\par
\hspace*{1.5em}D.3\quad \hyperref[app:ehr]{ICU lab acquisition under real-world missingness} \dotfill \pageref{app:ehr_results}\par
\hspace*{1.5em}D.4\quad \hyperref[app:diagnostics]{Empirical Diagnostics} \dotfill \pageref{app:diagnostics}\par
\endgroup

\newpage
\section{Theoretical Insights and Proofs}
\label{app:proofs}

\subsection{Identification Of Target Policy Value}
\label{app:identification_proof}
The AFA acquisition process induces an
additional set of endogenous variables: the action sequence $\underline A_T$, the
sequentially revealed features $\underline X_T$. We model the
target policy as a \emph{stochastic policy intervention}: the submodel
$M_{\operatorname{do}(\underline A_T \sim \pi_\alpha)}$ replaces the structural equation for
each action $A_t$ with a draw from $\pi_\alpha(A_t \mid S_t)$, where $S_t = (\underline X_{t}, \underline A_{t-1})$ and $S_0 = (X_0)$. We adopt a similar identification and estimation strategy described by \cite{von2023evaluationstatic}.

We restate the assumptions defined in \Cref{sec:setup}, as well as additional assumptions for completeness.

\begin{assumption}[Positivity]
\label{app_assumption:positivity}
    If $\mathbb{P}_{\pi_\alpha}(\tau) > 0$, then $\pi_\beta(R \geq R_T \mid X_o) > 0$, where $R_T$ denotes the set of acquisition actions $\underline A_T$ in the trajectory $\tau$.
\end{assumption}
\vspace{-5mm}
\begin{assumption}[Missing at Random]
\label{app_assumption:mar}
$X^{(1)} \ci R \mid X_o$, where $X_o$ denotes the subset of features always observed, i.e. $X^{(1)} = [X_o, X_u]$
\end{assumption}
\vspace{-5mm}
\begin{assumption}[No Direct Effect]
    The measurement of any feature $A_j$ does not affect any underlying features $X^{(1)}$ or the target variable $Y$.
\end{assumption}

We define a new distribution $\mathbb{P}_{\mathrm{sim}}$ induced by $M_{\operatorname{do}(\underline A_T \sim \pi_{\mathrm{sim}})}$, where each acquisition $A_t$ is drawn from the simulator defined as follows.

\textbf{The blocked simulator.} We define a simulation policy $\pi_{\mathrm{sim}}$ as a projection of the target policy $\pi_\alpha$ onto the support of the retrospectively available features. Let $r \in \{0, 1\}^d$ denote the realized binary missingness indicator for a given historical sample. By explicitly conditioning on $r$, the simulation policy strictly blocks the acquisition of unobserved features by setting their selection probability to zero, and renormalizes the probabilities of the remaining available actions:
\begin{align}
\pi_{\mathrm{sim}}(a_t \mid s_t, r) = \frac{\pi_\alpha(a_t \mid s_t) \cdot r_{a_t}}{\sum_{a' \in \mathcal{A}} \pi_\alpha(a' \mid s_t) \cdot r_{a'}}
\end{align}
The resulting joint simulated distribution $\mathbb{P}_{\mathrm{sim}} = \mathbb{P}_{\mathrm{sim}}(\underline X_{T}, \underline A_{T-1}, X_o, R, Y)$ induced by the simulator is given by:
\begin{align*}
&\mathbb{P}_{\mathrm{sim}}=\mathbb{P}(X_o, Y)\pi_\beta(R \mid X_o)\prod_{t=0}^{T-1} \pi_{\mathrm{sim}}(A_t \mid \underline X_t, R) \mathbb{P}(X_{t+1} \mid \underline X_t, \underline A_t, X_o, Y)
\end{align*}
Because the blocked simulation policy $\pi_{\mathrm{sim}}'$ only acquires available features (those indices $j$ with $R_j = 1$), we do not require any generative modeling of counterfactual feature values that are missing, and we can directly read off the values from $X_j^*$.

\textbf{Claim (identification).} Under no direct effect (NDE), static features, 
missingness at random (MAR), and positivity, the target is identified: there is an identifying distribution $\mathbb{Q}$, whose factors are
functionals of the observed-data law $\Pobs(X,R,Y)$, such that the
functional
\[
\bar G(\Pobs) := \mathbb{E}_{\mathbb{Q}}\!\left[w_T(\tau)C(\tau)\right]
\]
satisfies $G(\Ptar) = \bar G(\Pobs)$ and \begin{align*}
    w_T(\tau) = \frac{\mathbb I[R \geq R_T]\prod_{t'=0}^T\pi_\alpha(A_{t'} \mid S_{t'})}{\pi_\beta(R \geq R_T\mid X_o)\prod_{t'=0}^T\pi_{\mathrm{sim}}(A_{t'} \mid S_{t'}, R)},
\end{align*}
\textbf{Proof.}
Let $R_T \in \{0,1\}^d$ denote the indicator vector of final acquired
features (the d-dimensional vector version of $A_0,\dots,A_{T-1}$). 
We begin by introducing notation for the target quantity. Since
$\Ptar$ is itself the $\operatorname{do}(R=\mathbf 1)$ submodel of $M$ (the full law
where all features are retrospectively acquired), applying the additional policy
intervention on the acquisition actions composes the two interventions:
\[
\Ptar^{\,\operatorname{do}(\underline A \sim \pi_\alpha)}(C(\tau))
= \mathbb{P}_{M}^{\,\operatorname{do}(R=\mathbf 1,\ \underline A \sim \pi_\alpha)}(C(\tau)),
\]
the post-intervention distribution over the total return $C(\tau)$ obtained from $M$ by (i)
fixing $R=\mathbf 1$ so every requested feature is revealed from $X^{(1)}$, and
(ii) replacing the structural equation for each action $A_t$, $t=1,\dots,T-1$, with
a draw from $\pi_\alpha(A_t \mid \underline X_{t-1}, \underline A_{t-1})$.

We can write the expected return under this post-intervention target distribution:
\begin{align*}
\mathbb{E}_{\Ptar}[C(\tau)]
&= \sum_{\tau, x^{(1)}, y} C(\tau) p(x^{(1)}, y) \bigg(\prod_{t=0}^{T-1}\pi_\alpha(a_t \mid s_t) \mathbb{P}(x_{t+1} \mid \underline x_{t}, a_{t}, x^{(1)},  y)\bigg) \\
&= \sum_{\tau, x^{(1)}, y} C(\tau) \left( \sum_r p(x^{(1)}, y, r) \frac{\mathbb{I}[r \geq r_T]}{p(r \geq r_T \mid x^{(1)}, y)} \right) \bigg(\prod_{t=0}^{T-1}\pi_\alpha(a_t \mid s_t) \mathbb{P}(x_{t+1} \mid \dots )\bigg) \\
&= \sum_{\tau, x^{(1)}, y, r} C(\tau) \frac{\mathbb{I}[r \geq r_T] p(x^{(1)}, y, r)}{p(r \geq r_T \mid x_o)} \bigg(\prod_{t=0}^{T-1}\pi_\alpha(a_t \mid s_t) \mathbb{P}(x_{t+1} \mid \underline x_{t}, a_{t}, x^{(1)},  y)\bigg) \\
&= \sum_{\tau, x_o, y, r} C(\tau) \frac{\mathbb{I}[r \geq r_T] p(x_o, y, r)}{p(r \geq r_T \mid x_o)} \bigg(\prod_{t=0}^{T-1}\pi_\alpha(a_t \mid s_t) \mathbb{P}(x_{t+1} \mid \underline x_{t}, a_{t}, x_o,  y)\bigg)
\end{align*}
The first equality uses NDE assumption, where $\operatorname{do}(\underline A \sim \pi_\alpha)$ has no direct causal effect on the true underlying state $X^{(1)}$ or the outcome $Y$.

The second step requires the positivity assumption (\Cref{app_assumption:positivity}) to use the identity
$$ p(x^{(1)}, y) = \sum_r p(x^{(1)}, y, r) \frac{\mathbb{I}[r \geq r_T]}{p(r \geq r_T \mid x^{(1)}, y)}.$$
The third step requires the MAR assumption (\Cref{app_assumption:mar})
$$p(r \geq r_T \mid x^{(1)}, y) =  p(r \geq r_T \mid x_o).$$
The last step replaces $x^{(1)}$ with $x_o$, since we do not need to consider the full counterfactual distribution given $\mathbb I[r \geq r_T]$. In other words, since $x^{(1)} = [x_o, x_u]$ and we have that $X_{t+1} \ci X_u \mid X_o, Y$, $\mathbb{P}(x_{t+1} \mid \underline x_{t}, a_{t}, x_o, x_u, y) = \mathbb{P}(x_{t+1} \mid \underline x_{t}, a_{t}, x_o, y)$ and then we marginalize out $x_u$ via $\sum_{x_u} p(x_o, x_u, y,r) = p(x_o, y, r)$.

Since the simulation policy induces the following density over $(x_o, y, r, \underline x_T, \underline a_{T-1})$
$$p_{\mathrm{sim}} = p(x_o, y, r)\prod_{t=0}^{T-1}\pi_{\mathrm{sim}}(a_t \mid s_t, r) \mathbb{P}(x_{t+1} \mid \underline x_{t}, a_{t}, x_o,  y).$$
We can compute the identifying functional and estimator through importance weighting:
\begin{align*}
    &\mathbb{E}_{\Ptar}[C(\tau)] \\ &= \sum_{\tau, x_o, y, r} C(\tau) \frac{\mathbb{I}[r \geq r_T] p(x_o, y, r)}{p(r \geq r_T \mid x_o)p(x_o, y, r)} \bigg(\prod_{t=0}^{T-1}\frac{\pi_\alpha(a_t \mid s_t) \mathbb{P}(x_{t+1} \mid \underline x_{t}, a_{t}, x_o,  y)}{\pi_{\mathrm{sim}}(a_t \mid s_t, r)\mathbb{P}(x_{t+1} \mid \underline 
    x_{t}, a_{t}, x_o,  y)}\bigg)p_{\mathrm{sim}} \\
    &= \sum_{\tau, x_o, y, r} C(\tau) \frac{\mathbb{I}[r \geq r_T]}{p(r \geq r_T \mid x_o)} \bigg(\prod_{t=0}^{T-1}\frac{\pi_\alpha(a_t \mid s_t)}{\pi_{\mathrm{sim}}(a_t \mid s_t, r)}\bigg)p_{\mathrm{sim}} \\
    &= \sum_{\tau, x_o, y, r} w(\tau)C(\tau) p_{\mathrm{sim}} \\
    &= \mathbb{E}_{\mathbb{P}_{\mathrm{sim}}}[w_T(\tau)C(\tau)]
\end{align*}

\subsection{Extension to Regression}
\label{app:regression}
Our framework naturally extends to regression settings where the target variable is continuous, $Y~\in~\mathbb R$. While the main text measures uncertainty using predictive entropy for discrete outcomes, for regression it is standard to quantify uncertainty using predictive variance. 

During policy evaluation in the regression setting, we seek to estimate the expected return of the target policy, which we define as the pointwise variance reduction marginalized over the population's trajectory distribution:
\begin{align}
\label{eq:objective_regression}
G_{\mathrm{reg}}(\mathbb{P}_{\mathrm{tar}}) = 
\underbrace{\mathrm{Var}(Y \mid S_0, \mathbb{P}_{\mathrm{tar}}) - \mathrm{Var}(Y \mid S_T^{(\pi_\alpha)}, \mathbb{P}_{\mathrm{tar}}) \vphantom{\mathbb{E}_{\tau \sim \mathbb{P}_{\pi_\alpha}}\left[\sum_{t=0}^T c(A_t)\right]}}_{\text{Expected Variance Reduction}} - 
\underbrace{\mathbb{E}_{\tau \sim \mathbb{P}_{\pi_\alpha}}\left[\sum_{t=0}^T c(A_t)\right]}_{\text{Deterministic Costs}}.
\end{align}
Let the true conditional distribution $\tilde{\mathbb{P}}$ be parameterized by $\theta_{s_T}$ (for example, dictating the mean and variance of a Gaussian). Analogous to the entropy decomposition in the main text, we can decompose the total predictive variance of $Y$ after observing the dataset $z_{1:n}$ using the law of total variance:
\begin{align*}   
\label{eq:variance_decomp}
\mathrm{Var}(Y \mid s_T, z_{1:n}) &= \underbrace{\mathbb{E}\left[\mathrm{Var}(Y \mid s_T, \tilde{\mathbb P}) \mid z_{1:n} \right]}_{\text{Aleatoric uncertainty}~U_a(s_T)}  + \underbrace{\mathrm{Var}\left(\mathbb{E}[Y \mid s_T, \tilde{\mathbb P}] \mid z_{1:n} \right)}_{\text{Epistemic uncertainty}~U_e(s_T)}. 
% &= U_a(s_T) + U_e(s_T).
\end{align*}
In this continuous formulation, epistemic uncertainty $U_e$ is captured by the variance of the predicted mean under the posterior distribution of our parameters. It represents the uncertainty about the true parameter which resolves to zero as we collect more data. Conversely, the aleatoric uncertainty $U_a$ is the expected variance over the posterior distribution $\tilde{\mathbb{P}} \mid z_{1:n}$.
% , $U_a = \mathbb{E}\left[\mathrm{Var}(Y \mid s_T, \tilde{\mathbb P}) \mid z_{1:n} \right]$.

\subsection{Regret}
\label{app:regret}
Let $\tilde{\mathbb P}\mid z_{1:n}$ denote the posterior over the data-generating distribution. For a state $s$, the true negative entropy $-H(Y\mid s,\mathbb P^*)$ is unknown. Given the data, its best (Bayes-optimal) estimate is
its posterior expectation, which is the negative aleatoric uncertainty
$-U_a(s)$, where 
\begin{align*}
U_a(s) = \mathbb{E}\bigl[H(Y\mid s,\tilde{\mathbb P})
\mid z_{1:n}, s\bigr].
\end{align*}
Writing
$U(S_T^{(\pi)}) = \mathbb{E}\bigl[H(Y\mid S_T^{(\pi)},\tilde{\mathbb P})
\mid z_{1:n}\bigr]$ for the expectation
over the final-state distribution of $\pi$, we take the target objective
to be
\[
J(\pi_\alpha) = -\,U_a\bigl(S_T^{(\pi_\alpha)}\bigr).
\]
The surrogate objective instead uses the entropy of the posterior
predictive $P(Y\mid S_T, z_{1:n})$:
\[
\hat J(\pi_\alpha) = -\,H\bigl(Y\mid S_T^{(\pi_\alpha)}, z_{1:n}\bigr).
\]
By the decomposition
$H(Y\mid s, z_{1:n}) = U_a(s) + U_e(s)$, the two objectives differ by
exactly the epistemic uncertainty,
\[
\hat J(\pi) = J(\pi) - U_e\bigl(S_T^{(\pi)}\bigr),
\]
so the surrogate penalizes policies that end in epistemically uncertain
states.

Let $\pi^*$ and $\hat\pi$ maximize $J$ and $\hat J$, respectively, and
define the regret $J(\pi^*) - J(\hat\pi) \ge 0$.
Since $\hat J(\hat\pi) \ge \hat J(\pi^*)$ by construction,
\begin{equation}
\label{eq:regret}
0 \;\le\; J(\pi^*) - J(\hat\pi)
  \;\le\; U_e\bigl(S_T^{(\pi^*)}\bigr) - U_e\bigl(S_T^{(\hat\pi)}\bigr).
\end{equation}
Regret therefore arises only when the optimal policy leads to states with
more epistemic uncertainty than those reached by $\hat\pi$. Since $U_e(s)$ shrinks as $s$ is better covered by the offline data (e.g.\
$U_e(s) = O(1/n(s))$ under a Beta--Bernoulli model where $n(s)$ is the state visitation count), this happens when the optimal policy relies on states that are rare in $z_{1:n}$. This motivates selecting policies with $U_a$ rather than the predictive entropy.

\newpage

\section{Algorithms For Obtaining Posteriors}
\label{app:posterior_algorithms}

\subsection{The Bayesian Bootstrap}
\label{app:bayesian_bootstrap}
The Bayesian Bootstrap \citep{rubin1981bayesian} provides a nonparametric posterior over the empirical measure. Specifically, for a dataset of size $n$, we first draw a vector of probability weights from a uniform Dirichlet distribution, $w \sim \text{Dirichlet}(1, \dots, 1) \in \Delta^{n-1}$, and use it to parameterize a categorical distribution over the data indices. Intuitively, this Dirichlet weighting is directly equivalent to forward sampling within the martingale posterior framework. Simulating the unobserved future data conditional on the empirical support follows a predictive Pólya urn scheme, and the limit converges exactly to this Dirichlet distribution. By sampling $N-n$ indices with replacement from this distribution, we generate a single forward-sampled realization of the empirical measure.

\subsection{Martingale Posterior via Copulas}
\label{app:martingale_copula}
The sequential copula updates are used to instantiate the one-step predictive sequence for the marginale posterior, since it mathematically satisfies the \emph{martingale property} \citep{fongmartingale}. Intuitively,  the mean of the posterior samples is exactly equal to the initial $P_0(y)$, yielding unbiasedness $\mathbb{E}[P_\infty(y) \mid z_{1:n}] = P_0(y)$ (i.e. the posterior is centered around $P_0(y)$). Formally, the property stipulates that for every measurable set $B \subseteq \mathcal Z$ and $i \geq 0$, we have 
\begin{equation}
    \label{eq:martingale}
\mathbb{E}[P_{i+1}(B) \mid z_{1:i}] = P_i(B).
\end{equation}
Recall our goal is estimating the entropy for the state $s_T$ given by $H(Y \mid s_T)$. We consider a martingale posterior approach that combines a PFN (used at step 0) and copulas (used at steps~$>~1$) for the conditional $Y \mid X^*, R$ \citep{nagler2025uncertainty, melnychuk2026frequentist}, while the covariates $X^*, R$ are supplied by the Bayesian Bootstrap. In other words, we initialize $P_0(y \mid s_T, z_{1:n})$ using a PFN, and perform the following steps to obtain the MP updates for the predictive distribution of a query state instance $s_T$ for $j \in \{1, ..., N\}$ steps:

\begin{enumerate}
    \item Sample pseudo-observation $x_i', y_i'=(x^*, r, y)_i \sim P_i(X^*, R, Y).$
    \item Compute $v_i=P_{i}(y_{i}')=P_i(y_i' \mid x_i', z_{1:n}).$
    \item Obtain $P_{i}(y)=P_i(y \mid s_T, z_{1:n}).$
    \item Compute learning rate $\alpha_{i+1}(x_i', s_T) = \frac{\alpha_i\prod_{j=1}^d c_\rho(\Phi(x_i'^j), \Phi(s_T^j))}{1 - \alpha_i+\alpha_i\prod_{j=1}^d c_\rho(\Phi(x_i'^j), \Phi(s_T^j))}$ where $\alpha_i = (2-\frac{1}{n+i})\frac{1}{n+i+1}$.
    \item Compute $P_{i+1}(y \mid s_T, z_{1:n}) = \big( 1-\alpha_{i+1}(x_i', s_T) + \alpha_{i+1}(x_i', s_T)d_\rho(P_{i}(y), v_i)\big)P_{i}(y).$
\end{enumerate}

Given a correlation parameter $\rho$, the bivariate Gaussian copula is given by 
$$c_\rho(u, v) = \frac{\mathcal{N}_2(\Phi^{-1}(u), \Phi^{-1}(v) \mid 0, 1,  \rho)}{\mathcal{N}(\Phi^{-1}(u) \mid 0, 1)\mathcal{N}(\Phi^{-1}(v) \mid 0, 1)},$$
where inputs $u$ and $v$ are values between $[0,1]$. When computing the learning rate in step 4, instead of taking the product over all features $\prod_{j=1}^d c_\rho(\Phi(x_i'^j), \Phi(s_T^j))$, we only take the product over $j$ when both $x'^j$ and $s_T^j$ is observed, i.e. $j \in \{j_{obs}\}$. 

For the binary target variable in step 5, we use the Beta-Bernoulli bivariate copula which is given by \begin{align}
    d_\rho(u, v) \;=\;
\begin{cases}
  \displaystyle
  1-\rho+\rho\,\frac{\text{min}(u, v)}{uv},
    & y = y_j',\\[3ex]
  \displaystyle
  1-\rho+\rho\,\frac{u-\text{min}(u, 1-v)}
                    {uv},
    & y \neq y_{j}'.
\end{cases}
\end{align}

The copula method has a hyperparameter $\rho$ that controls the smoothness and volatility of the MP updates, with $\rho \rightarrow 0$ leading to independent updates with the copula density $d_\rho(u, v) \rightarrow 1$. We fix this value to 0.5 for our experiments. 

The choice of the learning rate $\alpha_i = (2-\frac{1}{n+i})\frac{1}{n+i+1}$ originates from \cite{fongmartingale}. 

\subsection{Predictive CLT}
\label{app:clt}
We consider the predictive CLT approach from \cite{fortini2026principled}. Here, we provide only a brief outline of the framework and its underlying assumptions, deferring detailed proofs and derivations to the original work. We target the conditional probability of $Y=1$ given the final acquisition state $s_T$: $m(s_T) := P(Y = 1 \mid s_T)$.

The predictive CLT establishes the following asymptotic normality result for $m_n(x)$ for some $x$:
\begin{align}
    P\left(\sqrt{n}(m_n(x) - m_\infty(x)) \mid Z_{1:n}\right) \rightarrow \mathcal{N}_m(0, V(x)) \quad P\text{-a.s.}
\end{align}
For asymptotic normality to hold, we require conditions on the volatility of the one-step predictive probability updates $\Delta_k(x) := m_k(x) - m_{k-1}(x)$ (given by Theorem 5.3 in \cite{fortini2026principled}). The increments are split into a martingale difference component and a conditional drift/bias component:
$$\Delta_k(x) = \zeta_k + \xi_k$$
where the martingale difference is the centered fluctuation term
$$\zeta_k := \Delta_k(x) - \mathbb{E}[\Delta_k(x) \mid Z_{1:k-1}],$$
and the directional drift term is
$$\xi_k := \mathbb{E}[\Delta_k(x) \mid Z_{1:k-1}].$$
Using this decomposition, the predictive error $m_n(x) - m_\infty(x)$ rewrites as
\begin{align*}
    m_n(x) - m_\infty(x) = \underbrace{\sum_{k \ge n+1} \zeta_k}_{\text{Martingale Fluctuation}} + \underbrace{\sum_{k \ge n+1} \xi_k}_{\text{Cumulative Bias}}
\end{align*}
% The martingale fluctuation component supplies a total variance equal to
% $$\text{Var}\left(\sum_{k \ge n+1} \zeta_k\right) = \sum_{k \ge n+1} \text{Var}(\zeta_k)$$
% because $\mathbb{E}[\zeta_k \mid Z_{1:k-1}] = 0$ ensures that $\zeta_k$ is a sequence with pairwise orthogonal increments.
In our setting, we have a finite set of test evaluation pairs $\{(x_i, y_i)\}_{i=1}^m$ where each $x_i = (s_T)_i$ corresponds to some partially observed state of covariates with missingness. We first require the following conditions to hold. The first two decay conditions ensures that the cumulative bias term and its contribution to the variance vanishes.

\emph{Dominated Drift Decay}: With $L_1$ domination (i.e. there exists a single non-negative random variable $Z$ with finite expected value such that $|U_n| \le Z$ almost surely for all $n$), 

$$\sqrt{n} \sum_{k > n} \mathbb{E}(\Delta_k(x_i) \mid Z_{1:k-1}) \rightarrow 0 \quad P\text{-a.s.}$$
$$\sqrt{n} \sup_{k \ge n} |\mathbb{E}(\Delta_k(x_i) \mid Z_{1:k-1})| \rightarrow 0 \quad P\text{-a.s.}$$

\emph{Squared Drift Decay}: The cumulative squared conditional drift vanishes at rate $n$
$$n\sum_{k \ge n} \mathbb{E}(\Delta_k(x_i) \mid Z_{1:k-1})^2 \to 0 \quad P\text{-a.s.}$$
\emph{Residual Volatility Limit}: $R_n(x) := n \sum_{k \ge n+1} \Delta_k(x) \Delta_k(x)^\top$ converges $P$-a.s. to a positive-definite random covariance matrix $V(x)$

\begin{remark}
    The original paper \citep{fortini2026principled} did not formally verify whether these theoretical conditions hold for a PFN, relying instead on empirical diagnostics. Furthermore, those diagnostics did not evaluate settings where $z_{1:n} = \{x, r, y\}_{1:n}$ contains missing features. To address this, we provide diagnostics demonstrating that reasonable variance estimates can still be obtained even when queries and contexts are only partially observed.
\end{remark}

\paragraph{Moment-matching for entropy estimation.}
We also employ moment-matching to convert the CLT mean $g_n(x)$ and covariance $V_n / n$, which parameterizes a Normal distribution, to a Beta approximation. We construct two Beta approximations, one which uses only the marginal (the \textbf{marginal} estimator in \cite{fortini2026principled}) and one which uses the entire covariance (the \textbf{CLT} estimator). 

We first clip the variance to $\sigma_n^2(x) < g_n(x)(1 - g_n(x))$. We determine the parameters $\alpha_n = \alpha_n(x)$ and $\beta_n = \beta_n(x)$ of the approximating Beta distribution $G \sim \text{Beta}(\alpha_n, \beta_n)$ by matching its first two moments to the CLT mean and the clipped variance:

$$\mathbb{E}[G] = \frac{\alpha_n}{\alpha_n + \beta_n} = g_n(x)$$
$$\text{Var}(G) = \frac{\alpha_n \beta_n}{(\alpha_n + \beta_n)^2(\alpha_n + \beta_n + 1)} = \sigma_n^2(x)$$

To incorporate the correlation structure over $m$ query points in the Beta distribution, we use a gaussian copula with beta margins.
$$R_n(x_i, x_j) = \frac{V_n(x_i, x_j)}{\sqrt{V_n(x_i, x_i)V_n(x_j, x_j)}}$$

$R_n = L_nL_n^T$ is decomposed via a symmetric eigendecomposition and negative eigenvalues are clipped and renormalized so each marginal variance is $1$. Then for each posterior sample $k$, we obtain a sample $Z^{(k)} \sim \mathcal N(0, I_m)$ and we obtain a standard normal sample with the correlation structure injected through $L_n$:
$$W^{(k)} = L_n Z^{(k)}$$
Then we obtain the final sampled prediction by passing the standard normal sample through the standard Normal CDF to map to the $\text{Unif}(0,1)$, and then passing that through the inverse Beta CDF $F^{-1}$ \citep{sklar1973random}: 
$$\theta^{(k)}(x_i)=F_{\alpha_n(x_i),\beta_n(x_i)}^{-1}(\Phi(W^{(k)}(x_i)))$$
This enables each $\theta^{(k)}(x_i) \sim \text{Beta}(\alpha_n(x_i), \beta_n(x_i))$ exactly by construction, but the samples across the queries are correlated across $m$ samples within a posterior draw $k$ as the probability integral transform (PIT) quantity $\Phi(W^{(k)}(x_i))$ captures the correlation between samples.

% \subsection{Evaluation}

% \textbf{Path stability}: The stability of the functional $G(P_N)$ as the rollout length $N$ increases is the primary diagnostic for the existence of a stable limit. We monitor this convergence by tracking the expected L1-norm
%     \begin{align*}
%         \mathbb{E}_{P_N} \left[ \frac{1}{p} \|G(P_n) - G(P_N)\|_1 \right].
%     \end{align*}
%      where $p$ is the parameter dimension. We check for the existence of a stable asymptote, i.e. if the TabPFN produces a consistent law, these trajectories should stabilize at a non-zero constant as $N$ grows. Systematic drift or divergent paths would indicate that $G(P_\infty)$ is ill-defined and the resulting finite-N posterior should be treated cautiously.

\newpage
\section{Experiment Details}
\label{app:experiments}

\subsection{Synthetic Experiments}
\label{app:synthetic}
\paragraph{Tasks.} We construct binary classification tasks from three DGP classes. All three classes share the following structure:

\begin{center}
\begin{tabular}{llll}
\toprule
Variable & Dimensionality & Meaning \\
\midrule
$x_o$    & $5$ & baseline features, never missing \\
$x_u$ & $5$ & lab features, acquirable features that are potentially missing \\
$d$      & $10$ & total feature dimension \\
$C$      & $2$ & classes (binary $y$) \\
\bottomrule
\end{tabular}
\end{center}

For each DGP sample, we draw various structural parameters to simulate diversity in the possible datasets, such as the prevalence of $Y=1$ and the number of informative features, along with their feature strengths.

\begin{center}\small
\begin{tabular}{lllp{4.4cm}}
\toprule
Range & Parameter \\
\midrule
$n_I \sim \mathcal{U}\{1,\dots,10\}$ & informative columns $\mathcal I$ \\
$a \sim \mathcal{U}(0.4,\,1.2)$ & class-separation strength \\
$\tau \sim \mathcal{U}(0,\,1)$ & log-sd of per-feature strengths (linear/gmm) \\
$\pi \sim \mathcal{U}(0.1,\,0.9)$ & marginal $P(y=1)$ \\
\bottomrule
\end{tabular}
\end{center}

For the Linear and GMM families, we allocate the signal across features. Given the task's spread $\tau$, per-feature strengths are drawn log-normally and renormalised so their squares average one,
\begin{equation*}
  \tilde s_j \sim \mathrm{LogNormal}(0,\tau^2), \qquad
  s_j \;=\; \tilde s_j \sqrt{\,n_{\mathcal{I}} \big/ \textstyle\sum_{i} \tilde s_i^2\,},
  \qquad j \in \mathcal{I},
  \label{eq:strengths}
\end{equation*}

\paragraph{Linear.} The linear DGP makes the features class-conditionally independent Gaussians with unit variance.
\begin{align*}
  y &\sim \mathrm{Bernoulli}(\pi), \\
  x \mid y &\sim \mathcal N(\mu_y,\, I_p), \qquad
  \mu_0 = 0, \qquad
  \mu_{1,j} = \begin{cases} a\, s_j\, \varepsilon_j, & j \in \mathcal{I},\\ 0, & \text{otherwise,}\end{cases}
  \label{eq:linear-gen} \\
\end{align*}
with independent random signs $\varepsilon_j \in \{-1,+1\}$. This leads to the following ground truth expression for the conditional distribution given any subset $x_S$
\begin{equation*}
  \operatorname{logit} p(y=1\mid x_S)
  \;=\; \log\frac{\pi}{1-\pi}
  \;+\; \sum_{j \in S} w_j\Big(x_j - \tfrac{\mu_{0,j}+\mu_{1,j}}{2}\Big),
  \qquad w = \mu_1 - \mu_0 .
  \label{eq:linear-post}
\end{equation*}
Noise columns have $w_j = 0$ and
so contribute nothing whether observed or not.

\paragraph{GMM.} The GMM family keeps closed-form tractability while
making the class-conditionals \emph{multimodal} and the boundary
\emph{nonlinear}. Each class is a mixture of $K$ diagonal Gaussians over the
informative columns. We first consider the additional structural parameters that are sampled per DGP:

\begin{center}\small
\begin{tabular}{lllp{4.9cm}}
\toprule
Variable & Range (per task) & Parameter \\
\midrule
$K$ & $\mathcal{U}\{2,3,4\}$ & Gaussian components \emph{per class} \\
$\varsigma$ & $\mathcal{U}(0.5,1.5)$ & standard deviation of the component center within a class \\
$\sigma^2$ & $\mathcal{U}(0.5,1.5)$ & diagonal variance of each component \\
$w_{y\cdot}$ & $\mathrm{Dirichlet}(2\cdot\mathbf{1}_K)$ & Mixture weights, drawn per class \\
\bottomrule
\end{tabular}
\end{center}

The DGP is then sampled as follows.
\begin{align*}
  y &\sim \mathrm{Bernoulli}(\pi), \quad
  c \mid y\sim \mathrm{Categorical}(w_{y}), \quad x_{\mathcal{I}} \mid y,c \sim \mathcal N \big(\mu_{y c},\, \sigma^2 I\big)
  \label{eq:gmm-gen}
\end{align*}
The Gaussian component means across the binary classes $y=0$ and $y=1$ are obtained through the following method, which introduces signal using a similar mechanism as the linear DGP:
\begin{equation*}
  \mu_{0 c} \;\sim\; \Normal\big(0,\; \varsigma^2 I_{n_{\mathcal{I}}}\big),
  \qquad
  \mu_{1 c} \;=\; \mu_{0 c} \;+\; a\,(s \odot \varepsilon),
  \qquad c = 1,\dots,K.
  \label{eq:gmm-centers}
\end{equation*}
The ground truth expression for the conditional distribution given any subset $x_s$ can be obtained as:
\begin{equation*}
  p(y=1 \mid x_S) \;\propto\; \pi
  \sum_{c=1}^{K} w_{1c} \prod_{j \in S \cap \mathcal{I}}
  \Normal\!\big(x_j;\, \mu_{1cj},\, \sigma^2\big).
  \label{eq:gmm-post}
\end{equation*}

\paragraph{BNN.} The BNN tasks are drawn from a random neural network:
\begin{align*}
  x &\sim \Normal(0, I_p), \\
  g(x) &= W_2^{\top} \tanh\!\big(W_1 x + b_1\big), \\
  f(x) &= \beta_0 + \lambda \cdot \frac{g(x) - m}{s}, \\
  y \mid x &\sim \mathrm{Bernoulli}\big(\sigma(f(x))\big).
  \label{eq:bnn-gen}
\end{align*}
The network is drawn once per task, with the informative-column structure
enforced by zeroing the non-informative input weights:
\begin{equation*}
  W_1[:,\,\mathcal{I}] \sim \Normal\!\Big(0,\ \frac{1}{n_{\mathcal{I}}}\Big),
  \qquad W_1[:,\,\mathcal{I}^{c}] = 0,
  \qquad b_1 \sim \Normal(0,\,0.5^2),
  \qquad W_2 \sim \Normal\!\Big(0,\ \frac{1}{H}\Big),
  \label{eq:bnn-weights}
\end{equation*}
where $H=16$ is the width of the network. For a fully observed row the posterior is exactly $\sigmoid(f(x))$. For an observed subset $S$ we use Monte Carlo estimation to obtain a ground truth,
\begin{equation*}
  p(y=1 \mid x_S) \;=\; \Ex_{x_{S^{c}} \sim \Normal(0,I)}
  \big[\,\sigmoid\big(f(x_S, x_{S^{c}})\big)\,\big],
  \label{eq:bnn-post}
\end{equation*}

\paragraph{Missingness mechanism.}
One mechanism is shared by pretraining and every synthetic evaluation. Lab $j$ of row $i$ is dropped independently using the standardized z-score $z_i$ for each sample's always-observed subset $x_o$: 
\begin{equation}
  p_{ij} \;=\; \sigma \big(\,\text{shift} + o_j + \beta_j^{\top} z_i \,\big)
  \label{eq:mar}
\end{equation}
with per-lab logit offsets $o_j \sim \Normal(0,\,1)$ (how much the
labs differ in how often they go missing) and slopes
$\beta_j \sim \Normal(0,\,1/|X_o|)$ on the standardised baseline, so each lab's missingness depends linearly on the always-observed subset $X_o$ with logit standard deviation $\approx$ \emph{strength} (zero gives MCAR). \texttt{Shift} is set during evaluation when we want the data to reach a certain global missingness rate. 

\paragraph{PFN Training.}
For meta-training a PFN model from scratch on each DGP class, we use the methodology in \cite{kobayashi2025learning}. We briefly summarize the approach below:

We instantiate a PFN using a transformer architecture with the following parameters:
\begin{center}\small
\begin{tabular}{lllp{4.4cm}}
\toprule
Component & Hyperparameter & Default \\
\midrule
layers & $6$ & pre-norm blocks (attention + GELU MLP) \\
width & $256$ & \\
heads & $4$ & head dim $64$ \\
MLP width & $512$ & \\
dropout & $0.0$ & \\
parameters & $\approx 3.17$M & \\
standardisation & on & Per-task mean/sd over \emph{observed} context cells \\
\bottomrule
\end{tabular}
\end{center}

The training procedure is as follows, where for each of $B$ items in a batch:
\begin{enumerate}
  \item \textbf{Draw a task.} Sample DGP parameters introduced above to instantiate the DGP.
  \item \textbf{Sample rows.} Draw $N$ fully observed and labeled rows $(X, y)$ from that task.
  \item \textbf{Build the context} Apply
    the linear-MAR mechanism of to the context labs with a per-task global rate drawn from $=[0,\,0.4]$  with a fresh per-task mechanism shape.
  \item \textbf{Build the queries.} Duplicate the same $N$ rows, apply a random
    permutation, hide their labels, and reveal a random subset of labs per row: the baseline is always observed and
    $k \sim \mathcal{U}\{0,\dots,k_{\max}\}$ labs are unmasked. Here $k=0$ is exactly the evaluation's baseline-only state and $k = 5$ its fully acquired state, so the
    model is trained on the whole acquisition lattice the reward is evaluated on.
  \item \textbf{Standardise} using statistics from the observed context samples, applied to both halves. The sequence is the concatenation $[\,\text{ctx}_1..\text{ctx}_N,\ \text{query}_1..\text{query}_N\,]$, of length $2N$.
  \item \textbf{Compute loss.} Compute the autoregressive loss on the queries, where each query at step $i$ conditions on all the context samples $<i-1$ using a permutation-invariant attention mask.
  \begin{equation}
  \mathcal{L} \;=\; -\frac{1}{BN}\sum_{b=1}^{B}\sum_{k=1}^{N}
  \log \hat p_{\theta}\big(y_{b,k}^{\mathrm{query}} \;\big|\;
  x_{b,k}^{\mathrm{query}},\ \mathcal{D}_{b,\,<\pi(k)}\big)
  \label{eq:loss}
\end{equation}
\end{enumerate}

The optimization parameters are DGP-family specific as shown below:
\begin{center}\small
\begin{tabular}{llll}
\toprule
Setting & Linear& GMM & BNN \\
\midrule
context rows $N$ & $300$ & $300$ & $500$ \\
sequence length & $600$ & $600$ & $1000$ \\
steps & $30{,}000$ & $50{,}000$ & $50{,}000$ \\
learning rate & $3\cdot10^{-4}$ & $1\cdot10^{-4}$ & $1\cdot10^{-4}$ \\
\midrule
\multicolumn{4}{l}{\emph{Shared across families}} \\
Batch size $B$ & \multicolumn{3}{l}{$16$ (one fresh DGP each)} \\
min context $m$  & \multicolumn{3}{l}{$2$} \\
optimiser & \multicolumn{3}{l}{AdamW, weight decay $0.01$} \\
schedule & \multicolumn{3}{l}{linear warmup over $5\%$ of steps, then cosine decay} \\
gradient clip & \multicolumn{3}{l}{$1.0$ (global norm)} \\
context missingness & \multicolumn{3}{l}{global rate $\sim \mathcal{U}(0,0.4)$ per task, linear MAR} \\
validation & \multicolumn{3}{l}{$16$ tasks, $200$ query rows, every $500$ steps} \\
\bottomrule
\end{tabular}
\end{center}

\paragraph{Evaluation: Policies and estimand.} We evaluate each estimator by comparing its predicted acquisition value and corresponding credible interval for a randomly selected feature against the ground-truth reference value. Evaluation data generating processes (DGPs) are sampled from the same parameter ranges as described above, paired with a linear missingness mechanism from \Cref{eq:mar}. To isolate the effect of epistemic uncertainty on estimator performance, we provide all methods with the true missingness probabilities (the oracle propensity model) instead of estimating the propensity from data. We use $n_{\text{eval}}=200$ held-out samples for each DGP to compute the policy value estimate.

\subsection{Tabular Benchmarks}
We use the following datasets for the tabular benchmarking experiment.
\label{app:tabular}
\begin{center}\small
\begin{tabular}{lrrrl}
\toprule
Key & OpenML id & Rows & $p$ & Positive class \\
\midrule
\cflag{adult}                 & 1590  & 48\,842  & 10 & \texttt{>50K} \\
\cflag{credit}              & 44089 & 16\,714  & 10 & \texttt{1} (delinquency) \\
\cflag{electricity}         & 151   & 45\,312  & 8  & \texttt{UP} \\
\cflag{magic\_telescope}      & 1120  & 19\,020  & 10 & \texttt{h} (hadron) \\
\cflag{higgs}               & 23512 & 98\,050  & 10 & \texttt{1} (signal) \\
\cflag{default\_credit\_card}  & 42477 & 30\,000  & 10 & \texttt{1} (default) \\
\cflag{covertype}           & 1596  & 581\,012 & 10 & \texttt{2} (lodgepole pine) \\
\cflag{california}        & 44090 & 20\,634  & 8  & \texttt{True} ($>$ median price) \\
\bottomrule
\end{tabular}
\end{center}

\paragraph{Evaluation: Policies and estimand.} We evaluate estimators by comparing their predicted acquisition values and credible intervals for a random feature set against the ground truth. Evaluation datasets of size $n \in \{1000, 2000\}$ are generated using an MCAR missingness mechanism. To isolate the effect of epistemic uncertainty, we supply all methods with the oracle propensity model (the true missingness probabilities) instead of estimating it empirically. Each sampled evaluation dataset is split into two halves: the first acts as context to fit the TabICL model, and the second is used to construct the value estimate.

\subsection{ICU lab acquisition under real-world missingness}
\label{app:ehr}
\paragraph{Cohorts.}
We use a public ICU dataset from PhysioNet. For both, each feature is the mean of a variable over a 48\,h window, and it is missing if the variable was never measured in that window.
\begin{itemize}
  % \item \emph{PhysioNet/CinC 2012} \citep{silva2012predicting}: 12{,}000 stays drawn from MIMIC-II. Each stay gives one row: the first 48\,h of the stay. We use 25 time-series variables plus age, gender and ICU type, for 28 features. The label is mortality between 48\,h and 72\,h after
  % admission. Patients who were censored (ex. mortality) inside the window are excluded.
  \item \emph{PhysioNet/CinC 2019} \citep{reyna2019early}: We use 27 hourly variables plus age and gender, for 29 features. The records
  cover the whole ICU stay, so one patient gives up to five overlapping windows
  $(24j,\,24j+48]$\,h, $j=0,\dots,4$. A window is used only if the stay covers
  it. The label is sepsis onset (the first positive \texttt{SepsisLabel}) within
  24\,h after the window ends. A window is dropped, along with all later windows
  of that patient, once onset has occurred by the end of the window.
\end{itemize}
Cohorts are restricted to an \emph{acute} population: the criteria are temperature
${>}38$ or ${<}36\,^\circ$C, heart rate ${>}100$ or ${<}60$, respiratory rate
${>}20$ or ${<}12$, systolic BP ${<}90$, MAP ${<}65$, and SpO$_2$ ${<}92\%$.

\paragraph{Panels and states.}
Features are grouped into clinical panels (\Cref{tab:ehr-panels}). A patient
state is a set of panels $S$, and $x^{(S)}$ is the feature vector with every
feature outside $S$ set to missing. Demographics and vitals form the baseline
$B$, which is always observed. The \emph{lab panels} $\mathcal{L}$ are what a policy can order.

\begin{table}[!ht]
\centering
\small
\caption{Panels in the ICU experiment. $B$ = always-observed baseline;
$\mathcal{L}$ = acquirable lab panels.}
\label{tab:ehr-panels}
\vspace{-0.1in}
\begin{tabular}{@{}ll@{}}
\toprule
 & PhysioNet 2019 \\
\midrule
$B$: demographics & Age, Gender \\
$B$: vitals & HR, O$_2$Sat, Temp, SBP, MAP, Resp \\
\midrule
$\mathcal{L}$: blood gas & HCO$_3$, FiO$_2$, pH, PaCO$_2$, SaO$_2$ \\
$\mathcal{L}$: metabolic & Ca, Cl, Creatinine, Glucose, Mg, PO$_4$, K \\
$\mathcal{L}$: lactate & Lactate \\
$\mathcal{L}$: CBC & WBC, Hgb, Platelets \\
$\mathcal{L}$: liver & AST, ALP, Bilirubin (total) \\
$\mathcal{L}$: coagulation & PTT, Fibrinogen \\
\bottomrule
\end{tabular}
\end{table}

% \begin{table}[ht]
% \centering
% \small
% \caption{Panels in the EHR experiment. $B$ = always-observed baseline;
% $\mathcal{L}$ = acquirable lab panels.}
% \label{tab:ehr-panels}
% \begin{tabular}{@{}lll@{}}
% \toprule
%  & PhysioNet 2012 & PhysioNet 2019 \\
% \midrule
% $B$: demographics & Age, Gender, Weight, ICUType & Age, Gender \\
% $B$: vitals & HR, Temp & HR, O$_2$Sat, Temp, SBP, MAP, Resp \\
% \midrule
% $\mathcal{L}$: blood gas & pH, PaCO$_2$, SaO$_2$ & HCO$_3$, FiO$_2$, pH, PaCO$_2$, SaO$_2$ \\
% $\mathcal{L}$: metabolic & Na, K, Creatinine, Glucose, HCO$_3$ & Ca, Cl, Creatinine, Glucose, Mg, PO$_4$, K \\
% $\mathcal{L}$: lactate & Lactate & Lactate \\
% $\mathcal{L}$: CBC & WBC, HCT, Platelets & WBC, Hgb, Platelets \\
% $\mathcal{L}$: liver & ALT, Bilirubin, Albumin & AST, ALP, Bilirubin (total) \\
% $\mathcal{L}$: coagulation & --- & PTT, Fibrinogen \\
% \bottomrule
% \end{tabular}
% \end{table}

\paragraph{Evaluation: Policies and estimand.}
For the single-step acquisition, each evaluation dataset draws an initial set of labs $I\subset\mathcal{L}$. Its
size is uniform on $\{0,\dots,3\}$, capped so that at least two labs remain.
The candidate policies are all one-step acquisitions from $B\cup I$: policy $k$
orders one lab $k\in\mathcal{L}\setminus I$ and ends in state $S_k=B\cup I\cup\{k\}$.

For the multistep policies, we sample binary escalation policies using the following procedure:
\begin{align*}
  (g_1,\dots,g_{|\mathcal{L}|}) &\sim \text{Uniform permutation of } \mathcal{L}
  && \text{(panel order)},\\
  \tau &\sim \text{Unif}(0.15,\,1.2) && \text{(one temperature per candidate)},\\
  b &\sim \text{Unif}\{2,\dots,|\mathcal{L}|\} && \text{(budget of \emph{new} panels)} \\
\end{align*}
Then at each step of the sequence, we sample a random feature $x_{chosen}$ among those acquired so far, and sample a probabilistic rule for acquiring the next feature in the panel order:
\begin{equation*}
  \pi_\alpha(a_t \mid s_t)
  \;=\;
  \sigma \left(
     \mathrm{sgn} \,\frac{x_{\text{chosen}} - \theta}{\tau}
  \right),
  \label{eq:ms-trigger}
\end{equation*}
where $\theta$ is a random threshold sampled from a standard normal and $\mathrm{sgn}$ is $\{-1, +1\}$ to determine whether high values or low values of $x_{chosen}$ will push towards acquisition. 

Each sampled evaluation dataset is split into two halves: the first acts as context to fit the TabICL model, and the second is used to construct the value estimate.

\subsection{Additional Experiment Details}
\label{app:experiment_params}

\subsubsection{Hyperparameters}
We provide additional configurations for each method in \Cref{tab:hyperparameters}.

\begin{table}[!ht]
\centering
\scriptsize
\setlength{\tabcolsep}{3.8pt}
\caption{Hyperparameters of the four experiments. $\{\cdot\}$ denotes a uniform draw \emph{per DGP}.}
\label{tab:hyperparameters}
\vspace{-0.1in}
\resizebox{\textwidth}{!}{
\begin{tabular}{lcccc}
\toprule
 & \multicolumn{1}{c}{Synthetic} & \multicolumn{1}{c}{Tabular} &
 \multicolumn{2}{c}{ICU (PhysioNet)} \\
\cmidrule(lr){2-2}\cmidrule(lr){3-3}\cmidrule(lr){4-5}
Hyperparameter & known truth & benchmark & one-step & multistep \\
\midrule
\multicolumn{5}{l}{\emph{Worlds and sweep}}\\
Tasks / families      & 3 families & 6 & 2 & 2 \\
Datasets Per Task        & 500 & 100 & 50 & 50 \\
$n$ & $(100,300) \text{ or } (100,500)$ & $(1000,2000)$ & $\{500,1000,1500\}$ & $\{500,1000\}$ \\
Eval patients            & $200$ & $n/2$ & $n/2$ & $n/2$ \\
Prevalence               & $(0.1,0.9)$ & native & $0.05$ & $0.05$ \\
Missingness              & MAR $(.1,.4)$ & MCAR $(.1,.4)$ & natural & natural \\
\addlinespace
\multicolumn{5}{l}{\emph{Policy class}}\\
Candidates per DGP       & $1$ & $5$ & all remaining & $5$ \\
Rollouts per patient $M$ & $1$ & $1$ & $1$ & $4$ \\
\addlinespace
\multicolumn{5}{l}{\emph{Posterior sampler}}\\
Posterior draws $K$      & $100$ & $200$ & $200$ & $200$ \\
Copula chain length $T$  & $500$ & $1000$ & $1000$ & $1000$ \\
Copula correlation $\rho$ & $0.5$ & $0.9$ & $0.9$ & $0.9$ \\
CLT permutations         & $1$ & $1$ & $1$ & $1$ \\
Bootstrap draws          & $200$ & $200$ & $200$ & $200$ \\
Backbone                 & PFN (pretrained) & \multicolumn{3}{c}{TabICL v2} \\
Ensemble size            &  & $4$ & $4$ & $4$ \\
\addlinespace
\multicolumn{5}{l}{\emph{Value stage}}\\
Weighting                & \multicolumn{4}{c}{H\'ajek (self-normalised)} \\
Propensity               & oracle & oracle & logistic & logistic \\
\bottomrule
\end{tabular}}
\end{table}

\subsubsection{Compute}
\label{app:computational_costs}
All experiments were run on a server with 4 NVIDIA H100 NVL GPUs, 2 Intel(R) Xeon(R) Platinum 8480+ CPUs (56 cores each) with 2Tb of memory.

\subsubsection{Source code}
\label{app:source_code}
\url{https://github.com/reAIM-Lab/BayesianAFA}

\clearpage
\section{Additional Results}
\label{app:results}

\subsection{Synthetic Experiments}
\label{app:synthetic_results}
\paragraph{Policy Rankings and Regret on Synthetic DGP families.}
We compute the improvement in pairwise accuracy and regret for each sampled evaluation dataset, across each global missingness level. We generate 500 evaluation datasets, and drop any datasets that contain all ties in the target feature entropy gain. 

\begin{table}[!ht]
\caption{Policy ranking of each variant against Total (plug-in) on the synthetic DGPs (Linear), one block per swept global missing rate: the mean paired difference (variant $-$ plug-in; higher is better for pairwise accuracy, lower for regret) over DGPs with its 95\% percentile bootstrap CI (DGPs resampled), and the one-sided Wilcoxon signed-rank $p$-value against the variant being no better than the baseline (higher pairwise accuracy, lower regret) (bold: $p < 0.05$.  $n$ = DGPs.)}
\label{tab:policy_ranking_vs_plugin_synthetic_linear}
\vspace{-0.1in}
% \begin{tiny}
\centering
\resizebox{\textwidth}{!}{
\begin{tabular}{llrcrrcr}
\toprule
 & & \multicolumn{3}{c}{Pairwise accuracy} & \multicolumn{3}{c}{Regret} \\
\cmidrule(lr){3-5}\cmidrule(lr){6-8}
Missing rate & Variant & $\Delta$ & 95\% CI & $p$ & $\Delta$ & 95\% CI & $p$ \\
\midrule
\multirow[t]{5}{*}{0 ($n=496$)} & Total (bootstrap) & $-0.005$ & $[-0.010, -0.001]$ & 0.993 & $+0.0000$ & $[-0.0006, +0.0005]$ & 0.718 \\
 & Bootstrap & $+0.004$ & $[+0.000, +0.009]$ & \textbf{0.027} & $-0.0001$ & $[-0.0007, +0.0004]$ & 0.241 \\
 & Marginal (clt) & $+0.007$ & $[+0.001, +0.014]$ & \textbf{0.006} & $-0.0001$ & $[-0.0003, +0.0001]$ & \textbf{0.016} \\
 & Copula & $-0.002$ & $[-0.004, +0.000]$ & 0.946 & $+0.0000$ & $[-0.0002, +0.0002]$ & 0.673 \\
 & CLT & $+0.005$ & $[-0.000, +0.012]$ & \textbf{0.013} & $-0.0002$ & $[-0.0005, +0.0002]$ & 0.157 \\
\midrule
\multirow[t]{5}{*}{0.1 ($n=496$)} & Total (bootstrap) & $-0.003$ & $[-0.007, +0.001]$ & 0.862 & $-0.0002$ & $[-0.0009, +0.0003]$ & 0.764 \\
 & Bootstrap & $+0.006$ & $[+0.001, +0.010]$ & \textbf{0.002} & $-0.0004$ & $[-0.0011, +0.0001]$ & 0.494 \\
 & Marginal (clt) & $+0.003$ & $[-0.001, +0.008]$ & 0.083 & $+0.0001$ & $[-0.0001, +0.0004]$ & 0.921 \\
 & Copula & $+0.001$ & $[-0.003, +0.004]$ & 0.250 & $-0.0000$ & $[-0.0005, +0.0004]$ & 0.934 \\
 & CLT & $+0.006$ & $[+0.001, +0.011]$ & \textbf{0.011} & $-0.0001$ & $[-0.0005, +0.0002]$ & 0.676 \\
\midrule
\multirow[t]{5}{*}{0.2 ($n=496$)} & Total (bootstrap) & $-0.004$ & $[-0.010, +0.001]$ & 0.772 & $-0.0001$ & $[-0.0004, +0.0002]$ & 0.779 \\
 & Bootstrap & $+0.005$ & $[-0.002, +0.010]$ & \textbf{$<$0.001} & $-0.0003$ & $[-0.0007, +0.0001]$ & 0.403 \\
 & Marginal (clt) & $+0.004$ & $[-0.003, +0.011]$ & \textbf{0.040} & $-0.0002$ & $[-0.0007, +0.0003]$ & 0.591 \\
 & Copula & $-0.006$ & $[-0.011, -0.001]$ & 0.968 & $+0.0001$ & $[-0.0000, +0.0004]$ & 0.650 \\
 & CLT & $+0.003$ & $[-0.004, +0.010]$ & 0.055 & $+0.0000$ & $[-0.0005, +0.0005]$ & 0.803 \\
\midrule
\multirow[t]{5}{*}{0.3 ($n=496$)} & Total (bootstrap) & $+0.000$ & $[-0.004, +0.005]$ & 0.665 & $+0.0002$ & $[-0.0004, +0.0008]$ & 0.958 \\
 & Bootstrap & $+0.007$ & $[+0.002, +0.012]$ & \textbf{0.013} & $+0.0001$ & $[-0.0005, +0.0006]$ & 0.936 \\
 & Marginal (clt) & $+0.010$ & $[+0.004, +0.016]$ & \textbf{$<$0.001} & $-0.0002$ & $[-0.0006, +0.0002]$ & 0.054 \\
 & Copula & $+0.001$ & $[-0.003, +0.004]$ & 0.367 & $-0.0001$ & $[-0.0004, +0.0001]$ & 0.497 \\
 & CLT & $+0.009$ & $[+0.003, +0.015]$ & \textbf{0.004} & $+0.0002$ & $[-0.0004, +0.0007]$ & 0.718 \\
\midrule
\multirow[t]{5}{*}{0.4 ($n=496$)} & Total (bootstrap) & $-0.000$ & $[-0.005, +0.004]$ & 0.442 & $+0.0001$ & $[-0.0005, +0.0007]$ & 0.676 \\
 & Bootstrap & $+0.004$ & $[-0.001, +0.009]$ & 0.100 & $+0.0000$ & $[-0.0004, +0.0006]$ & 0.395 \\
 & Marginal (clt) & $+0.007$ & $[-0.001, +0.014]$ & \textbf{0.041} & $-0.0002$ & $[-0.0012, +0.0009]$ & 0.497 \\
 & Copula & $-0.003$ & $[-0.007, -0.000]$ & 0.967 & $-0.0002$ & $[-0.0007, +0.0004]$ & 0.606 \\
 & CLT & $+0.004$ & $[-0.003, +0.012]$ & 0.165 & $+0.0005$ & $[-0.0008, +0.0019]$ & 0.690 \\
\bottomrule
\end{tabular}}
% \end{tiny}
\end{table}

\begin{table}[!ht]
\caption{Policy ranking of each variant against Total (plug-in) on the synthetic DGPs (GMM), one block per swept global missing rate: the mean paired difference (variant $-$ plug-in; higher is better for pairwise accuracy, lower for regret) over DGPs with its 95\% percentile bootstrap CI (DGPs resampled), and the one-sided Wilcoxon signed-rank $p$-value against the variant being no better than the baseline (higher pairwise accuracy, lower regret) (bold: $p < 0.05$.  $n$ = DGPs.)}
\label{tab:policy_ranking_vs_plugin_synthetic_gmm}
\vspace{-0.1in}
% \begin{tiny}
\centering
\resizebox{\textwidth}{!}{
\begin{tabular}{llrcrrcr}
\toprule
 & & \multicolumn{3}{c}{Pairwise accuracy} & \multicolumn{3}{c}{Regret} \\
\cmidrule(lr){3-5}\cmidrule(lr){6-8}
Missing rate & Variant & $\Delta$ & 95\% CI & $p$ & $\Delta$ & 95\% CI & $p$ \\
\midrule
\multirow[t]{5}{*}{0 ($n=499$)} & Total (bootstrap) & $+0.003$ & $[-0.002, +0.007]$ & 0.167 & $-0.0002$ & $[-0.0008, +0.0004]$ & 0.185 \\
 & Bootstrap & $+0.007$ & $[+0.002, +0.013]$ & \textbf{0.032} & $-0.0000$ & $[-0.0006, +0.0006]$ & 0.419 \\
 & Marginal (clt) & $-0.003$ & $[-0.008, +0.003]$ & 0.872 & $+0.0009$ & $[+0.0002, +0.0018]$ & 0.980 \\
 & Copula & $+0.004$ & $[+0.000, +0.007]$ & \textbf{0.008} & $+0.0001$ & $[-0.0004, +0.0006]$ & 0.371 \\
 & CLT & $-0.001$ & $[-0.006, +0.004]$ & 0.606 & $+0.0007$ & $[+0.0000, +0.0015]$ & 0.991 \\
\midrule
\multirow[t]{5}{*}{0.1 ($n=499$)} & Total (bootstrap) & $+0.002$ & $[-0.003, +0.007]$ & 0.107 & $+0.0000$ & $[-0.0002, +0.0003]$ & 0.398 \\
 & Bootstrap & $+0.010$ & $[+0.004, +0.018]$ & \textbf{0.004} & $-0.0001$ & $[-0.0003, +0.0002]$ & 0.126 \\
 & Marginal (clt) & $+0.004$ & $[-0.003, +0.011]$ & 0.222 & $+0.0002$ & $[-0.0002, +0.0005]$ & 0.242 \\
 & Copula & $+0.002$ & $[-0.002, +0.006]$ & 0.420 & $-0.0001$ & $[-0.0004, +0.0002]$ & 0.160 \\
 & CLT & $+0.004$ & $[-0.002, +0.011]$ & 0.145 & $+0.0001$ & $[-0.0002, +0.0005]$ & 0.103 \\
\midrule
\multirow[t]{5}{*}{0.2 ($n=499$)} & Total (bootstrap) & $-0.005$ & $[-0.010, +0.000]$ & 0.934 & $-0.0004$ & $[-0.0013, +0.0004]$ & 0.775 \\
 & Bootstrap & $+0.007$ & $[+0.002, +0.012]$ & \textbf{0.005} & $-0.0012$ & $[-0.0020, -0.0004]$ & \textbf{0.003} \\
 & Marginal (clt) & $+0.008$ & $[+0.003, +0.014]$ & \textbf{0.003} & $-0.0006$ & $[-0.0012, +0.0000]$ & \textbf{0.010} \\
 & Copula & $+0.004$ & $[+0.000, +0.009]$ & \textbf{0.029} & $-0.0002$ & $[-0.0008, +0.0003]$ & 0.409 \\
 & CLT & $+0.007$ & $[+0.001, +0.012]$ & \textbf{0.015} & $-0.0003$ & $[-0.0010, +0.0004]$ & 0.144 \\
\midrule
\multirow[t]{5}{*}{0.3 ($n=499$)} & Total (bootstrap) & $-0.003$ & $[-0.008, +0.002]$ & 0.835 & $-0.0003$ & $[-0.0014, +0.0006]$ & 0.340 \\
 & Bootstrap & $+0.008$ & $[+0.002, +0.014]$ & \textbf{0.023} & $-0.0004$ & $[-0.0016, +0.0006]$ & 0.187 \\
 & Marginal (clt) & $+0.010$ & $[+0.004, +0.017]$ & \textbf{$<$0.001} & $+0.0002$ & $[-0.0006, +0.0011]$ & 0.340 \\
 & Copula & $+0.002$ & $[-0.002, +0.006]$ & 0.175 & $+0.0002$ & $[-0.0005, +0.0008]$ & 0.313 \\
 & CLT & $+0.011$ & $[+0.005, +0.017]$ & \textbf{$<$0.001} & $+0.0000$ & $[-0.0009, +0.0009]$ & 0.172 \\
\midrule
\multirow[t]{5}{*}{0.4 ($n=499$)} & Total (bootstrap) & $-0.007$ & $[-0.013, -0.001]$ & 0.994 & $+0.0008$ & $[+0.0001, +0.0015]$ & 1.000 \\
 & Bootstrap & $+0.006$ & $[-0.000, +0.012]$ & 0.111 & $+0.0007$ & $[-0.0002, +0.0016]$ & 0.992 \\
 & Marginal (clt) & $+0.015$ & $[+0.008, +0.022]$ & \textbf{$<$0.001} & $+0.0005$ & $[-0.0006, +0.0018]$ & 0.691 \\
 & Copula & $+0.007$ & $[+0.002, +0.012]$ & \textbf{0.004} & $-0.0002$ & $[-0.0005, +0.0001]$ & 0.497 \\
 & CLT & $+0.015$ & $[+0.008, +0.022]$ & \textbf{$<$0.001} & $+0.0009$ & $[-0.0003, +0.0022]$ & 0.699 \\
\bottomrule
\end{tabular}}
% \end{tiny}
\end{table}

\begin{table}[!ht]
\caption{Policy ranking of each variant against Total (plug-in) on the synthetic DGPs (BNN), one block per swept global missing rate: the mean paired difference (variant $-$ plug-in; higher is better for pairwise accuracy, lower for regret) over DGPs with its 95\% percentile bootstrap CI (DGPs resampled), and the one-sided Wilcoxon signed-rank $p$-value against the variant being no better than the baseline (higher pairwise accuracy, lower regret) (bold: $p < 0.05$.  $n$ = DGPs.)}
\label{tab:policy_ranking_vs_plugin_synthetic_bnn}
\vspace{-0.1in}
% \begin{tiny}
\centering
\resizebox{\textwidth}{!}{
\begin{tabular}{llrcrrcr}
\toprule
 & & \multicolumn{3}{c}{Pairwise accuracy} & \multicolumn{3}{c}{Regret} \\
\cmidrule(lr){3-5}\cmidrule(lr){6-8}
Missing rate & Variant & $\Delta$ & 95\% CI & $p$ & $\Delta$ & 95\% CI & $p$ \\
\midrule
\multirow[t]{5}{*}{0 ($n=499$)} & Total (bootstrap) & $+0.002$ & $[-0.003, +0.006]$ & 0.122 & $-0.0001$ & $[-0.0006, +0.0004]$ & 0.143 \\
 & Bootstrap & $+0.001$ & $[-0.003, +0.005]$ & 0.213 & $+0.0002$ & $[-0.0002, +0.0007]$ & 0.185 \\
 & Marginal (clt) & $-0.002$ & $[-0.006, +0.002]$ & 0.783 & $-0.0001$ & $[-0.0008, +0.0003]$ & 0.499 \\
 & Copula & $+0.001$ & $[-0.001, +0.003]$ & 0.157 & $-0.0002$ & $[-0.0006, +0.0000]$ & 0.079 \\
 & CLT & $-0.001$ & $[-0.004, +0.003]$ & 0.630 & $-0.0001$ & $[-0.0008, +0.0004]$ & 0.240 \\
\midrule
\multirow[t]{5}{*}{0.1 ($n=499$)} & Total (bootstrap) & $+0.003$ & $[-0.002, +0.007]$ & 0.057 & $-0.0006$ & $[-0.0014, -0.0000]$ & 0.122 \\
 & Bootstrap & $+0.009$ & $[+0.004, +0.013]$ & \textbf{$<$0.001} & $-0.0006$ & $[-0.0012, -0.0000]$ & \textbf{0.022} \\
 & Marginal (clt) & $+0.004$ & $[-0.001, +0.009]$ & \textbf{0.046} & $-0.0001$ & $[-0.0006, +0.0005]$ & 0.498 \\
 & Copula & $-0.000$ & $[-0.003, +0.002]$ & 0.641 & $-0.0001$ & $[-0.0005, +0.0002]$ & 0.841 \\
 & CLT & $+0.004$ & $[-0.001, +0.009]$ & \textbf{0.040} & $-0.0002$ & $[-0.0007, +0.0004]$ & 0.114 \\
\midrule
\multirow[t]{5}{*}{0.2 ($n=499$)} & Total (bootstrap) & $-0.006$ & $[-0.011, -0.001]$ & 0.953 & $+0.0004$ & $[-0.0003, +0.0011]$ & 0.936 \\
 & Bootstrap & $+0.001$ & $[-0.004, +0.006]$ & 0.388 & $+0.0001$ & $[-0.0009, +0.0009]$ & 0.779 \\
 & Marginal (clt) & $+0.007$ & $[+0.002, +0.013]$ & \textbf{0.028} & $-0.0002$ & $[-0.0010, +0.0006]$ & 0.219 \\
 & Copula & $+0.001$ & $[-0.003, +0.004]$ & 0.369 & $+0.0002$ & $[-0.0003, +0.0008]$ & 0.502 \\
 & CLT & $+0.007$ & $[+0.001, +0.013]$ & 0.057 & $-0.0003$ & $[-0.0013, +0.0007]$ & 0.186 \\
\midrule
\multirow[t]{5}{*}{0.3 ($n=499$)} & Total (bootstrap) & $-0.007$ & $[-0.013, -0.002]$ & 0.987 & $+0.0003$ & $[-0.0006, +0.0012]$ & 0.936 \\
 & Bootstrap & $+0.008$ & $[+0.002, +0.013]$ & \textbf{0.004} & $-0.0004$ & $[-0.0013, +0.0005]$ & 0.416 \\
 & Marginal (clt) & $+0.021$ & $[+0.015, +0.027]$ & \textbf{$<$0.001} & $-0.0013$ & $[-0.0024, -0.0003]$ & \textbf{0.002} \\
 & Copula & $+0.002$ & $[-0.002, +0.006]$ & 0.066 & $-0.0002$ & $[-0.0012, +0.0005]$ & 0.499 \\
 & CLT & $+0.018$ & $[+0.012, +0.024]$ & \textbf{$<$0.001} & $-0.0015$ & $[-0.0029, -0.0003]$ & \textbf{0.004} \\
\midrule
\multirow[t]{5}{*}{0.4 ($n=499$)} & Total (bootstrap) & $-0.010$ & $[-0.015, -0.005]$ & 1.000 & $+0.0009$ & $[+0.0001, +0.0020]$ & 0.956 \\
 & Bootstrap & $-0.001$ & $[-0.005, +0.004]$ & 0.582 & $+0.0009$ & $[+0.0001, +0.0021]$ & 0.831 \\
 & Marginal (clt) & $+0.010$ & $[+0.004, +0.016]$ & \textbf{0.004} & $-0.0004$ & $[-0.0016, +0.0008]$ & 0.567 \\
 & Copula & $-0.000$ & $[-0.004, +0.004]$ & 0.323 & $-0.0003$ & $[-0.0012, +0.0006]$ & 0.603 \\
 & CLT & $+0.011$ & $[+0.005, +0.018]$ & \textbf{$<$0.001} & $-0.0004$ & $[-0.0017, +0.0010]$ & 0.345 \\
\bottomrule
\end{tabular}}
% \end{tiny}
\end{table}

\clearpage

\subsection{Tabular Benchmarks}
\label{app:tabular_results}
\paragraph{Value Estimation Bias and Coverage on OpenML Tabular Benchmarks.} Across subsampled datasets of $n \in [1000, 2000]$, we compute the value estimate for acquiring a set of features randomly selected among all possible candidate features in the dataset. The reference value is computed using the \textbf{Total (plug-in)} estimator with TabICL over $N = 10000$ samples from the same task. 

\begin{figure}[!ht]
    \centering
    \includegraphics[width=0.8\linewidth]{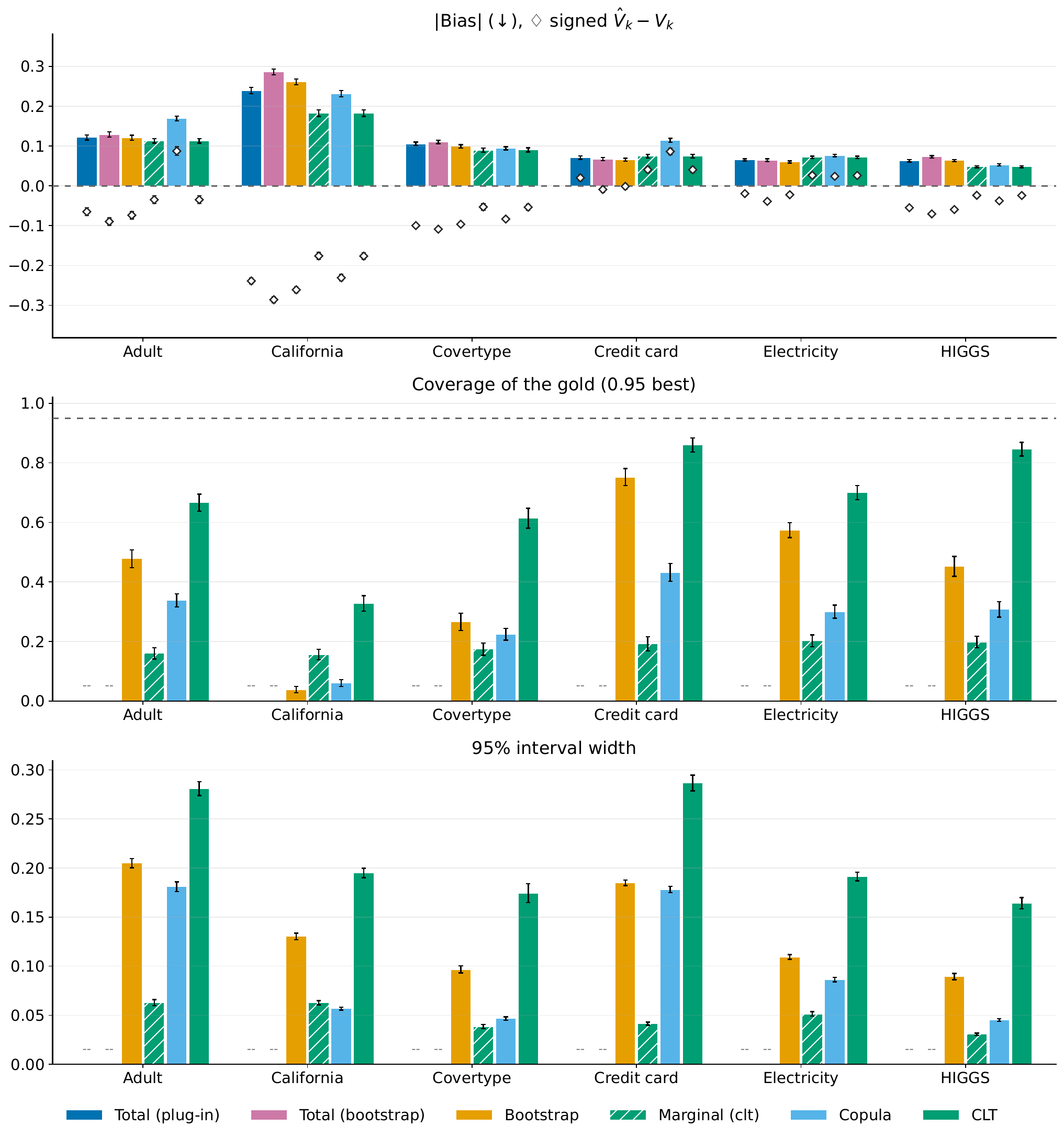}
    \caption{\textbf{Value Estimation Bias and Coverage on OpenML Tabular Benchmarks.} The mean bias (absolute and signed), coverage, and interval width and standard error is computed over 500 random single-step acquisition policies from 100 sampled datasets for each task.}
    \label{fig:tabular_bias}
\end{figure}

For the \textbf{Total (plug-in)} estimator, a large negative signed bias, together with an absolute bias of similar magnitude, indicates that the plug-in negative entropy computed on the subsampled data systematically underestimates the reference. Since the bias shrinks as more observations support a state, the gap relative to the large-sample reference is approximately $U_e^{(n)}(S_T) - U_e^{(N)}(S_T) > 0$, and it is largest when the final states are sparsely covered by the subsampled data, for example when acquired features take many distinct values. The magnitude of the bias thus theoretically reflects how poorly the subsampled data cover the final states reached, i.e. how difficult the final-state entropy is to \emph{estimate}. Therefore, tasks such as \textbf{California} and \textbf{Covertype} may contain a larger spread of possible distinct states (ex. many continuous features) relative to sample size.

In contrast, for \textbf{Credit Card} and \textbf{Electricity}, the
\textbf{Total (plug-in)} estimate shows only a small bias relative to the reference. One explanation is that the final states reached are well covered by the subsampled data, so $U_e^{(n)}(S_T)$ is small and the final-state entropy is recovered even at the reduced sample size. An alternative is overconfidence of the predictive model at the subsample
size: if the posterior predictive does not carry the epistemic uncertainty appropriate to $n$ observations, its entropy is too low,
which offsets the expected underestimation of the negative entropy.

\clearpage 
\paragraph{Policy Rankings and Regret on OpenML Tabular Benchmarks.} Across subsampled datasets of $n \in [1000, 2000]$, we compute the value estimate across \emph{5 candidate policies}, each acquiring a set of features randomly selected among all possible candidate features in the dataset. The reference values for each policy is computed using the \textbf{Total (plug-in)} estimator with the TabICL model over a larger $N = 10000$ set of samples from the same dataset. The reference ranking between the 5 candidate policies is then constructed using the reference value estimates.

\begin{figure}[!ht]
    \centering
    \includegraphics[width=\linewidth]{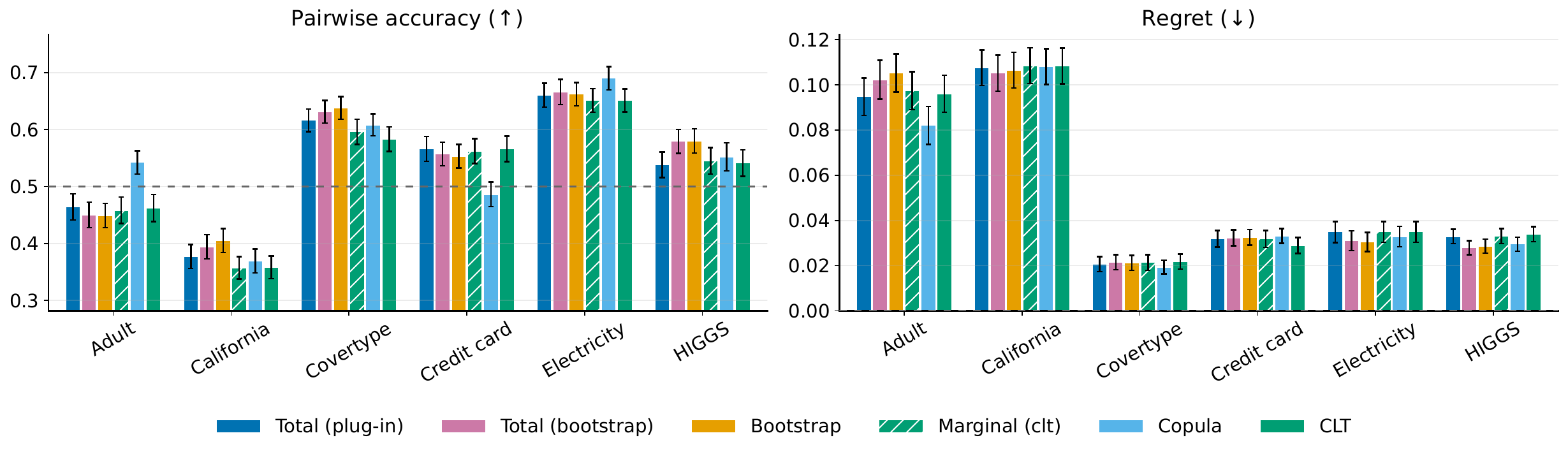}
    \caption{\textbf{Policy ranking and selection on OpenML tabular
    benchmarks.} For each task, 100 datasets are subsampled, and on each
    dataset 5 randomly generated candidate policies are evaluated with
    each estimator. \emph{Pairwise accuracy} is the fraction of policy
    pairs ordered consistently with the reference (0.5 corresponds to
    random ordering). \emph{Regret} is the reference value of the best
    candidate minus that of the policy selected by the estimator. Results
    are averaged over the 100 datasets.}
    \label{fig:tabular_ranking}
\end{figure}

On the tabular benchmarks, the bias-corrected estimators (bootstrap and BPI variants) do not yield statistically significant improvements in
policy selection over the plug-in estimators
(\Cref{fig:tabular_ranking}). All estimators attain similar pairwise accuracy and regret within each task, while the level around which they concentrate varies across tasks. Since the estimators share the same predictive entropy and differ only in
the epistemic correction, this pattern suggests that the ranking errors
are driven by a component common to all estimators, such as approximation
error of the underlying predictive model (TabICL), rather than by
the epistemic term that the correction targets. 

On \textbf{Adult} and \textbf{California}, however, the regret is large
and the pairwise accuracy falls below 0.5. Because random estimation
errors alone would drive pairwise accuracy toward 0.5 rather than below
it, this indicates a systematic error. A plausible mechanism is that better policies (as defined by the reference $J_N$) reach more specific and hence more sparsely covered final states in the subsampled data, where the negative entropy is underestimated most strongly (i.e. shrinks to the prior entropy), so that the estimation error reverses the ranking. A well calibrated $\hat U_e$ estimate from the bias-correction should theoretically remove this (as shown in the copula martingale posterior for \textbf{Adult}). We explore the reasons for this failure mode in the next section.

\clearpage
\paragraph{Correlation between posterior spread (estimated epistemic uncertainty) and value underestimation.} We verify if the policy-level value underestimation is predicted by the posterior spread of the value. We use the spread as a proxy for the epistemic correction $U_e(S_T)$. If the plug-in
underestimation is driven by epistemic uncertainty, policies with wider posteriors should have more negative errors, yielding a negative
association.

\begin{figure}[!ht]
    \centering
    \includegraphics[width=0.62\linewidth]{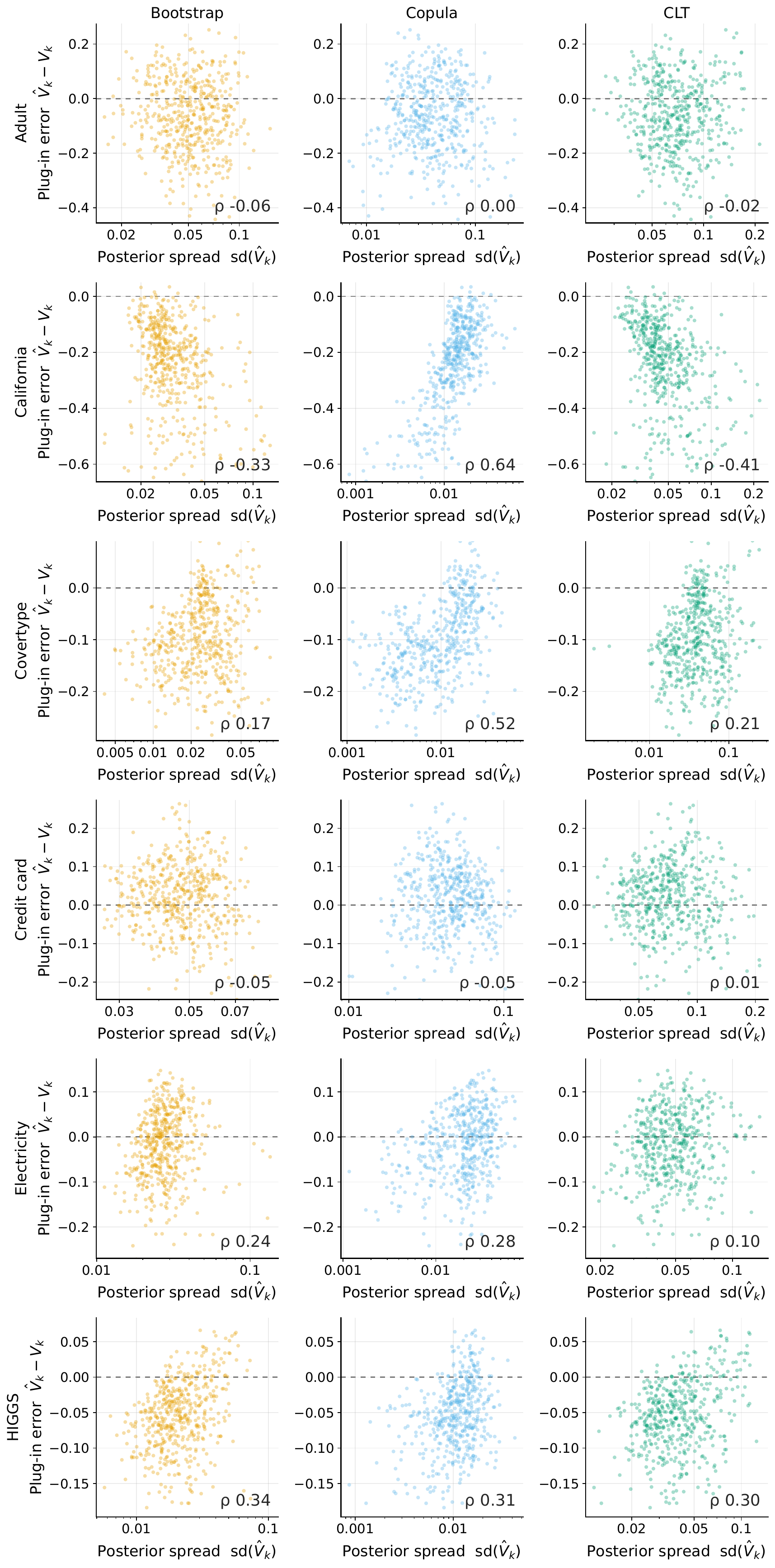}
    \caption{\textbf{Correlation between posterior spread (estimated epistemic uncertainty) and value underestimation.} Each point represents a value estimate from a candidate policy. Several tabular datasets show no such correlation, implying that the plug-in error is either small, indicating adequate coverage or the plug-in displays systematic overconfidence. Alternatively, error is driven by sources that the posterior spread does not reflect, such as approximation error of the predictive model.}
    \label{fig:tabular_spread}
\end{figure}

\clearpage
\subsection{ICU Prediction}
\label{app:ehr_results}
\paragraph{Single-Step and Multistep Acquisition Coverage and Width on Sepsis.} Across subsampled datasets of $n \in \{500, 1000, 1500\}$, we compute the value estimate for acquiring a set of features randomly selected among all possible candidate features in the dataset. The reference value is computed using the \textbf{Total (plug-in)} estimator with TabICL over  $N_{\text{sepsis}} = 10000$ samples from the same task. 

\begin{figure}[!ht]
    \centering
    \includegraphics[width=\linewidth]{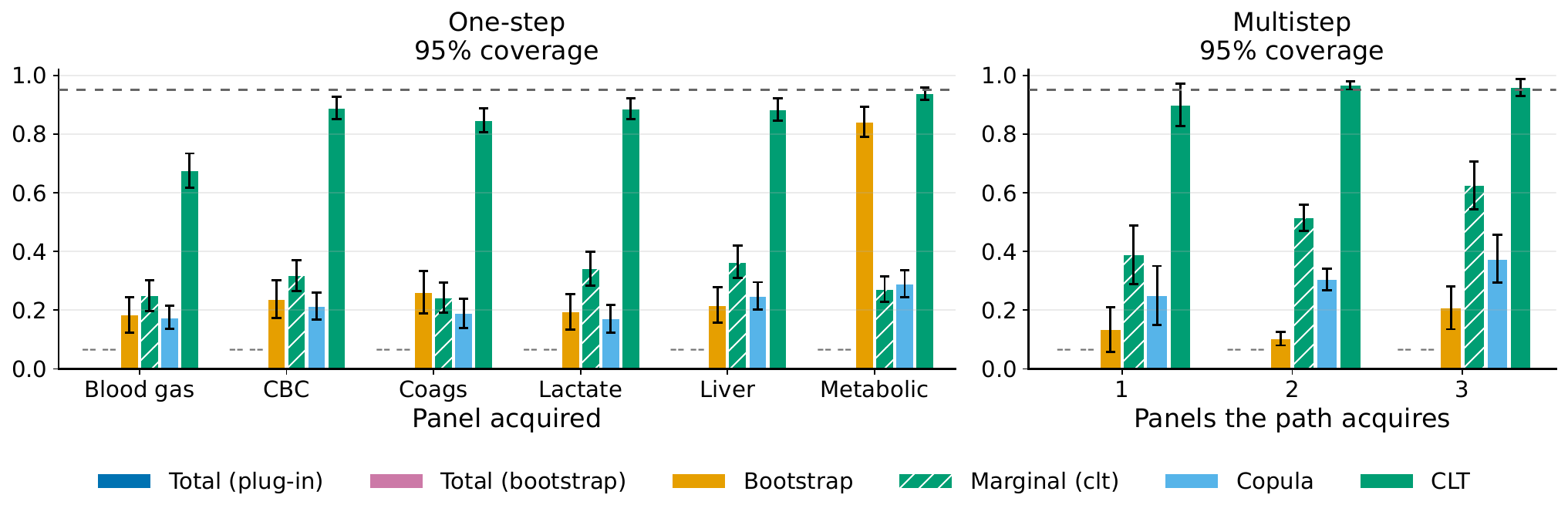}
    \caption{\textbf{Single-Step and Multistep Acquisition Coverage on Sepsis.} Coverage across each panel/feature group, averaged over 150 sampled evaluation datasets per task.}
    \label{fig:icu_coverage}
\end{figure}

\begin{figure}[!ht]
    \centering
    \includegraphics[width=\linewidth]{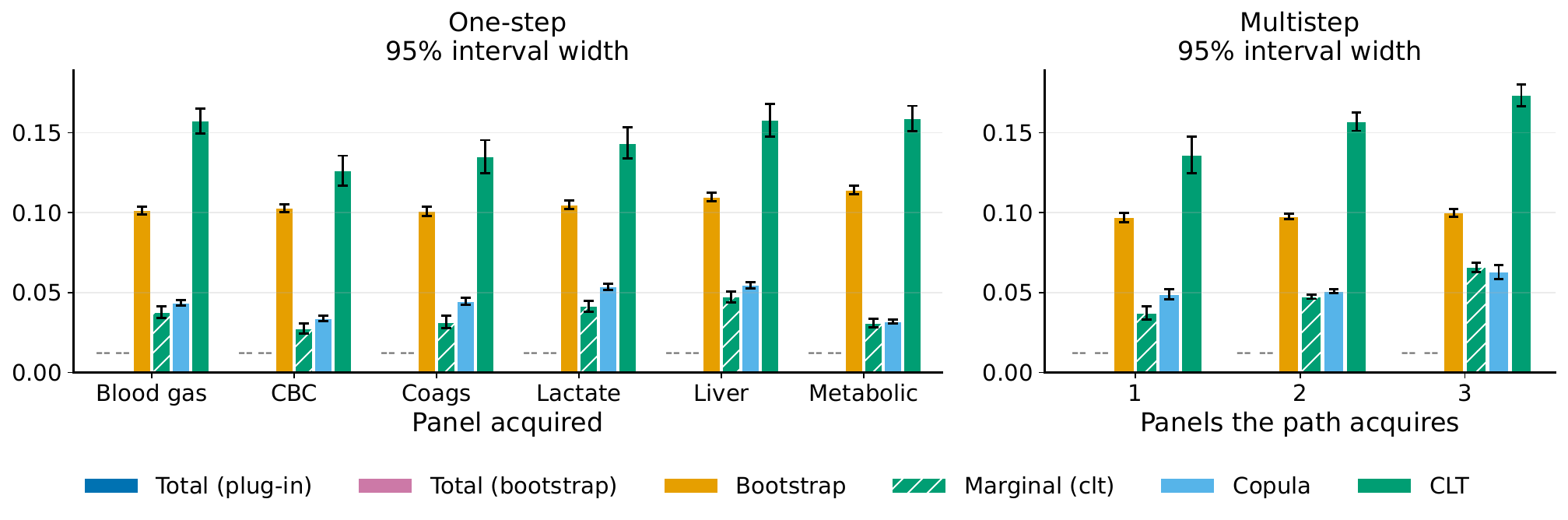}
    \caption{\textbf{Single-Step and Multistep Acquisition Interval Width on Sepsis.} Interval width across each panel/feature group, averaged over 150 sampled evaluation datasets per task.}
    \label{fig:icu_width}
\end{figure}

% In the \textbf{Mortality} task, the plug-in estimate shows no systematic underestimation of the negative entropy for some lab panel acquisitions (lactate, CBC), suggesting that the resulting final states may be adequately
% covered at the subsample size. For other panels, such as blood gas and metabolic, the plug-in systematically underestimates the negative entropy, consistent with epistemic uncertainty at sparsely covered final states,
% and the bias-corrected estimators reduce both the signed and the absolute bias. Because the full \textbf{Mortality} dataset is relatively small, the reference is itself a plug-in estimate that remains subject to epistemic
% bias at sparsely covered states. The measured bias therefore reflects only the difference $U_e^{(n)} - U_e^{(N)}$ and understates the epistemic bias of the plug-in estimate.

% The measured bias can also be reduced if the predictive model is overconfident at smaller context sizes. If the posterior predictive that
% TabICL produces from $n$ context samples does not carry the epistemic uncertainty appropriate to $n$ observations, its entropy is too low, which offsets the expected underestimation of the negative entropy. In
% Fig.~\ref{fig:ehr_sanity_check}, the plug-in and reference estimates differ only slightly on average (0.002 to 0.008). This is consistent
% with each of the explanations above (adequate coverage, epistemic bias in the reference, or overconfidence at the subsample size) but does not distinguish between them.

\clearpage 
\paragraph{Pairwise Accuracy and Regret on Sepsis.}
Across subsampled datasets of $n \in \{500, 1000\}$, we compute the value estimate across candidate policies as described in \Cref{app:ehr}.

\begin{figure}[!ht]
    \centering
    \includegraphics[width=0.9\linewidth]{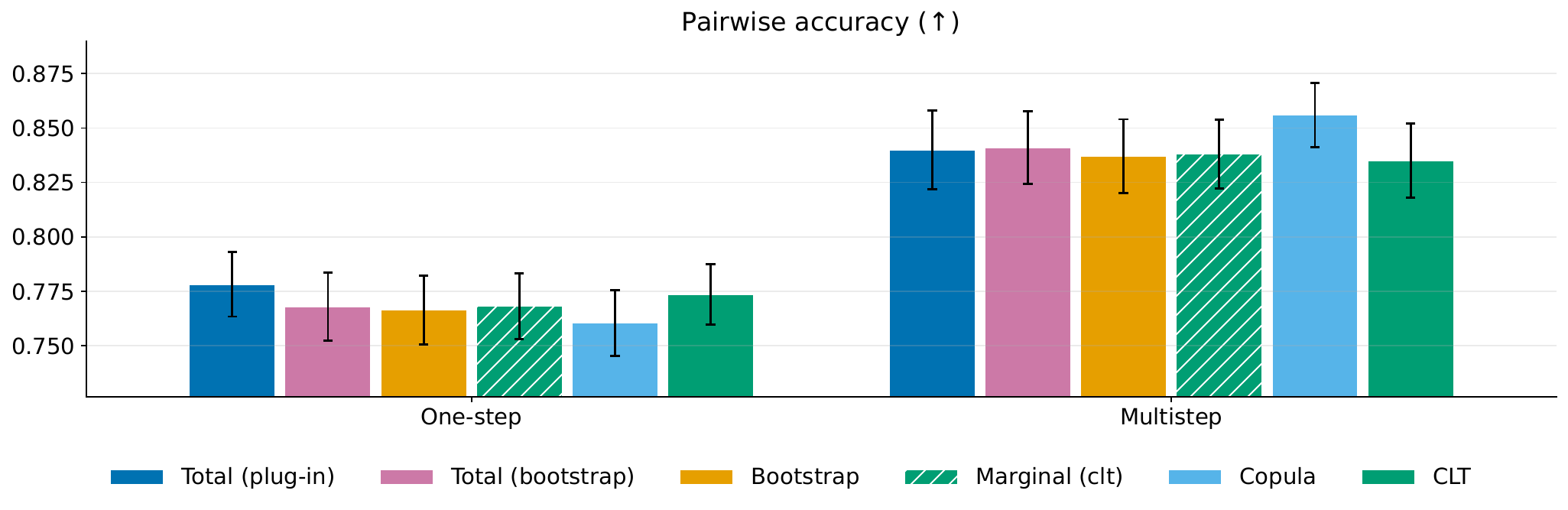}
    \caption{\textbf{Pairwise Accuracy on Sepsis.} Single-step acquisitions and multistep stochastic policies}
    \label{fig:icu_ranking}
\end{figure}

\begin{figure}[!ht]
    \centering
    \includegraphics[width=0.9\linewidth]{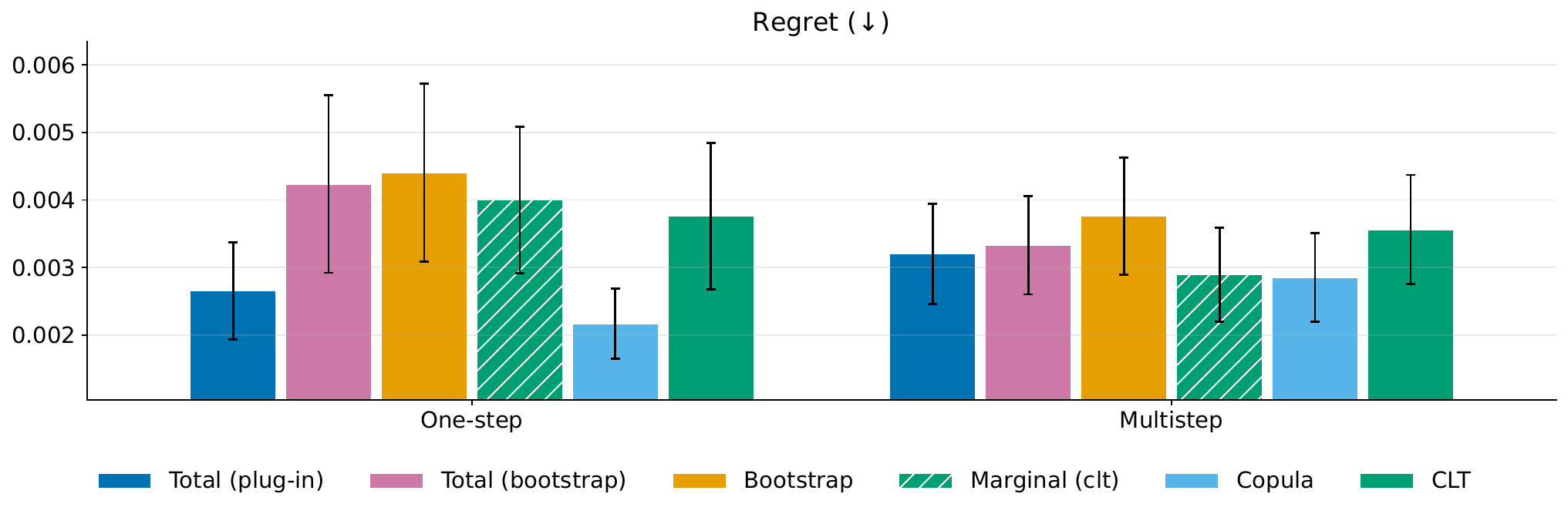}
    \caption{\textbf{Regret on Sepsis.} Single-step acquisitions and multistep stochastic policies}
    \label{fig:icu_ranking}
\end{figure}

The bias-corrected estimators (bootstrap and BPI variants) again do not yield statistically significant improvements in
policy selection over the plug-in estimators
(\Cref{fig:icu_ranking}). This pattern suggests that the ranking errors
are driven by a component common to all estimators, such as approximation
error of the underlying predictive model (TabICL), rather than by
the epistemic term that the correction targets. 

\paragraph{Spread of propensity estimates for single-step acquistiion across sample size on Sepsis.}
Across subsampled datasets of $n \in \{500, 1000, 1500\}$, we validate whether coverage is sufficient for single acquisition from retrospective missingness.

\begin{figure}[!ht]
    \centering
    \includegraphics[width=0.95\linewidth]{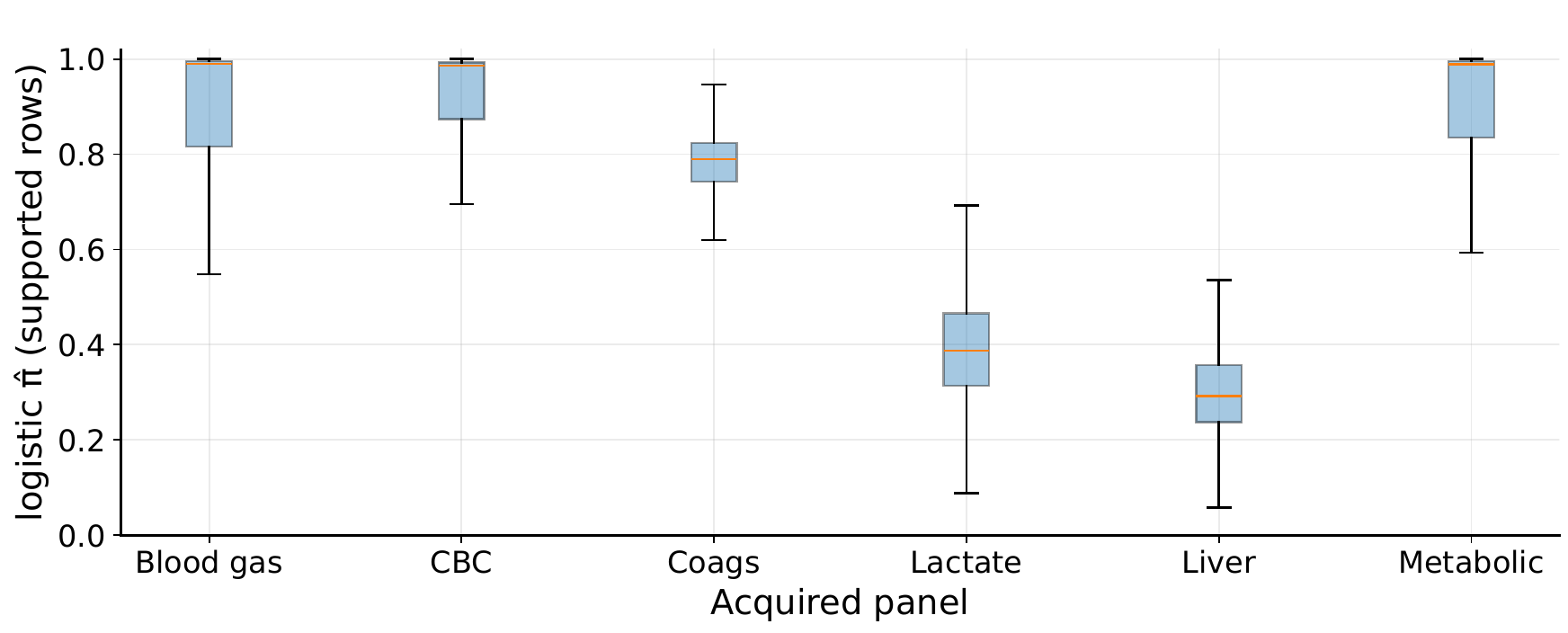}
    \caption{\textbf{Spread of propensity estimates for single-step acquistiion across sample size on Sepsis.}}
    \label{fig:ehr_propensity}
\end{figure}

\clearpage
\paragraph{Correlation between posterior spread (estimated epistemic uncertainty) and value underestimation.} We verify if the policy-level value underestimation is predicted by the posterior spread of the value. We use the spread as a proxy for the epistemic correction $U_e(S_T)$. If the plug-in
underestimation is driven by epistemic uncertainty, policies with wider posteriors should have more negative errors, yielding a negative
association.

\begin{figure}[!ht]
    \centering
    \includegraphics[width=\linewidth]{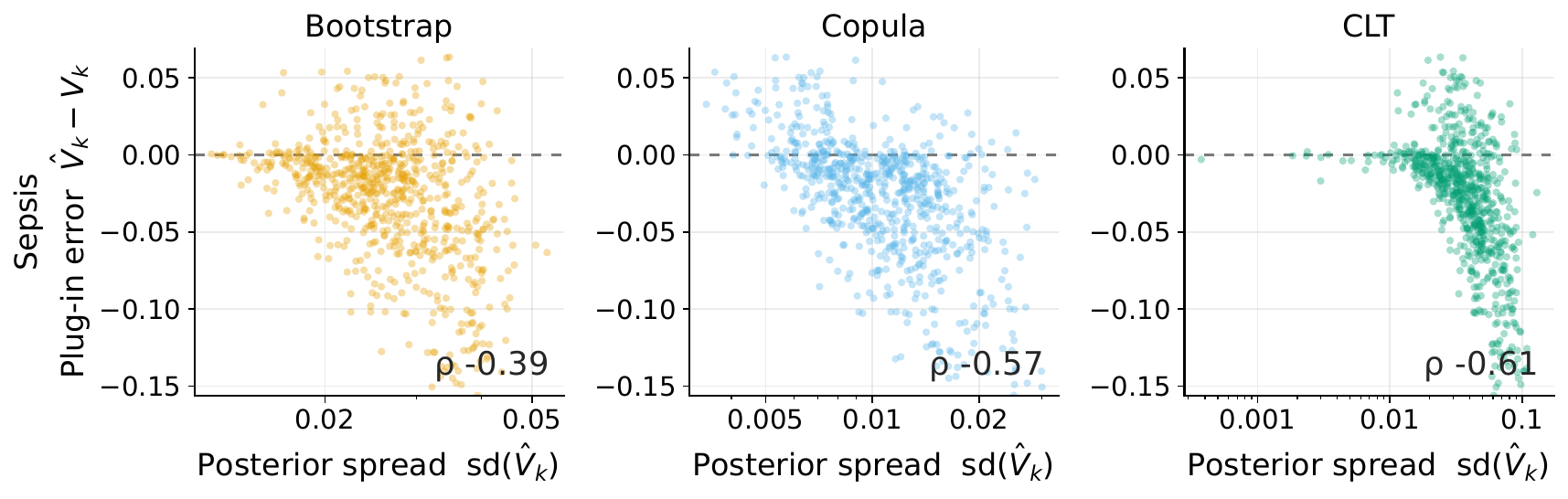}
    \caption{\textbf{Correlation between posterior spread (estimated epistemic uncertainty) and value underestimation.} Each point represents a value estimate from a candidate policy.}
    \label{fig:icu_spread}
\end{figure}

In \textbf{Sepsis}, the posterior spread is negatively correlated with the plug-in error (Spearman $\rho = -0.57/-0.61$) for copula and CLT respectively: policies whose final states are more sparsely covered are more strongly underestimated. The epistemic correction should therefore reduce the bias most for these policies. A correlation between spread and underestimation does not, however,
guarantee better policy rankings and pairwise accuracy. Rankings depend only on the differences in error between candidate policies evaluated on the same data, so the
correction improves them only if it varies sufficiently across these
policies relative to their value gaps.

\clearpage
\subsection{Empirical Diagnostics}
\label{app:diagnostics}
We first verify that TabICL generates calibrated PPDs $p = P(Y \mid s_T, z_{1:n})$, ensuring the predicted probabilities remain stable and converge toward the reference predictions $p_{\text{gold}} = P(Y \mid s_T, z_{1:N})$. We then confirm that each posterior estimation method behaves reasonably, as evidenced by improving coverage and shrinking interval widths as sample size increases. The coverage and width are computed sample-wise for each predicted probability $p = P(Y \mid s_T, z_{1:n})$, instead of the downstream population-level functionals.

\paragraph{Sample intervals of $p$}
We visualize sample posterior intervals for the pointwise predictions, verifying whether the variance is properly estimated. We take a random subset of patients for a randomly sampled evaluation setting. Monte Carlo estimation was used to obtain the posterior intervals for each method, and the reference predictions at $N$ as well as the plug-in predictions are also visualized.

\begin{figure}[!ht]
    \centering
    \includegraphics[width=0.95\linewidth]{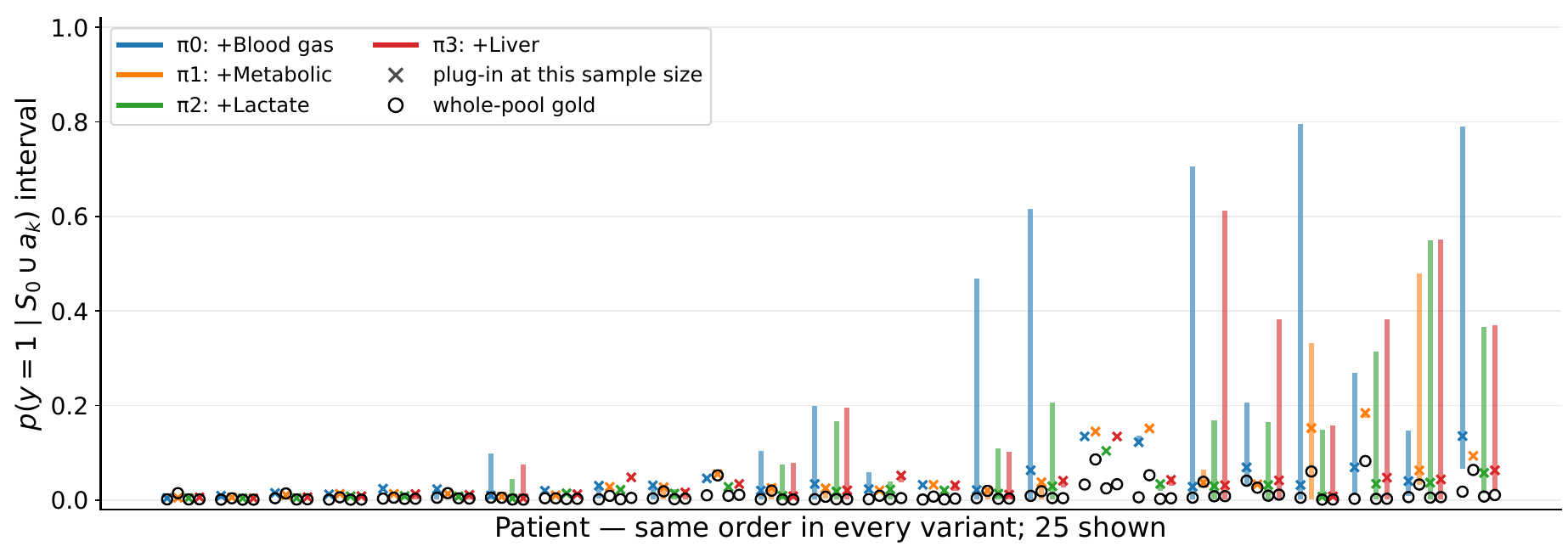}
    \caption{\textbf{Sample intervals of $p$ for Copula-based martingale posterior.}}
    \label{fig:intervals_copula}
\end{figure}

\begin{figure}[!ht]
    \centering
    \includegraphics[width=0.95\linewidth]{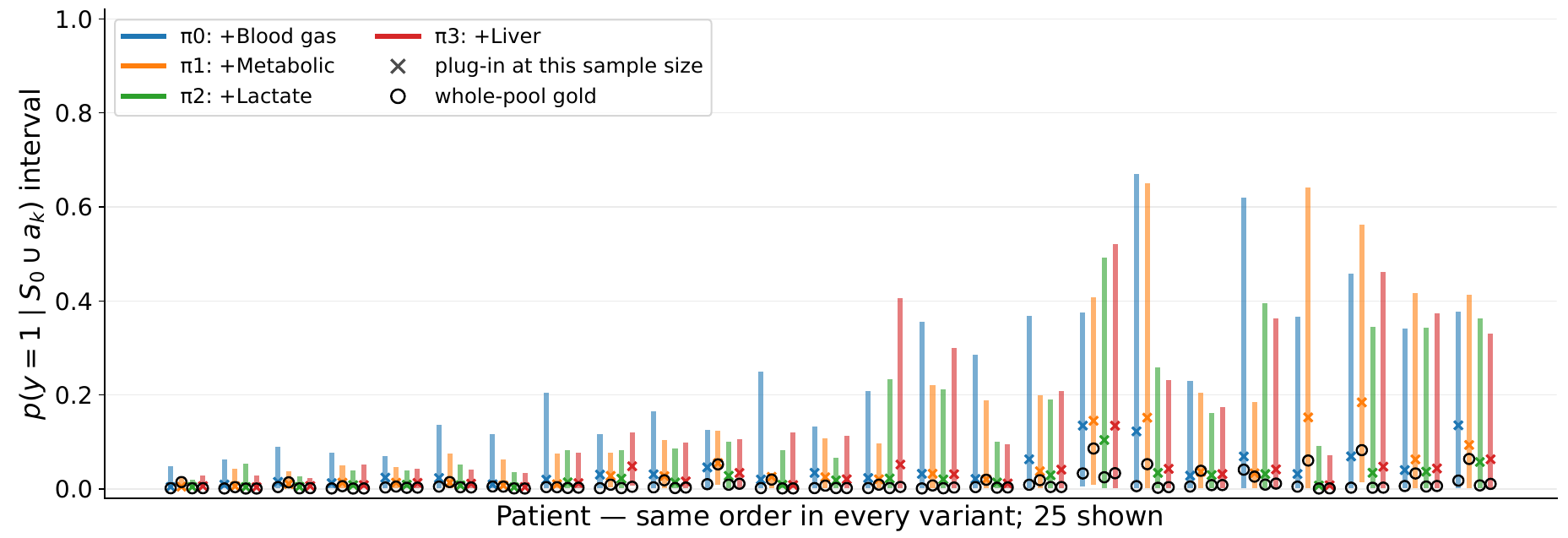}
    \caption{\textbf{Sample intervals of $p$ for predictive CLT posterior.}}
    \label{fig:intervals_clt}
\end{figure}

\clearpage
\paragraph{Posterior contraction for single-step acquisition across sample size on Tabular Benchmarks.}
Across subsampled datasets of $n \in [1000, 2000]$, we validate whether the resulting posterior samples for $P(Y \mid S_T)$ contract (reduce interval width as sample size increases). We split the evaluation dataset into 3 bins of equal size.

\begin{figure}[!ht]
    \centering
    \includegraphics[width=0.8\linewidth]{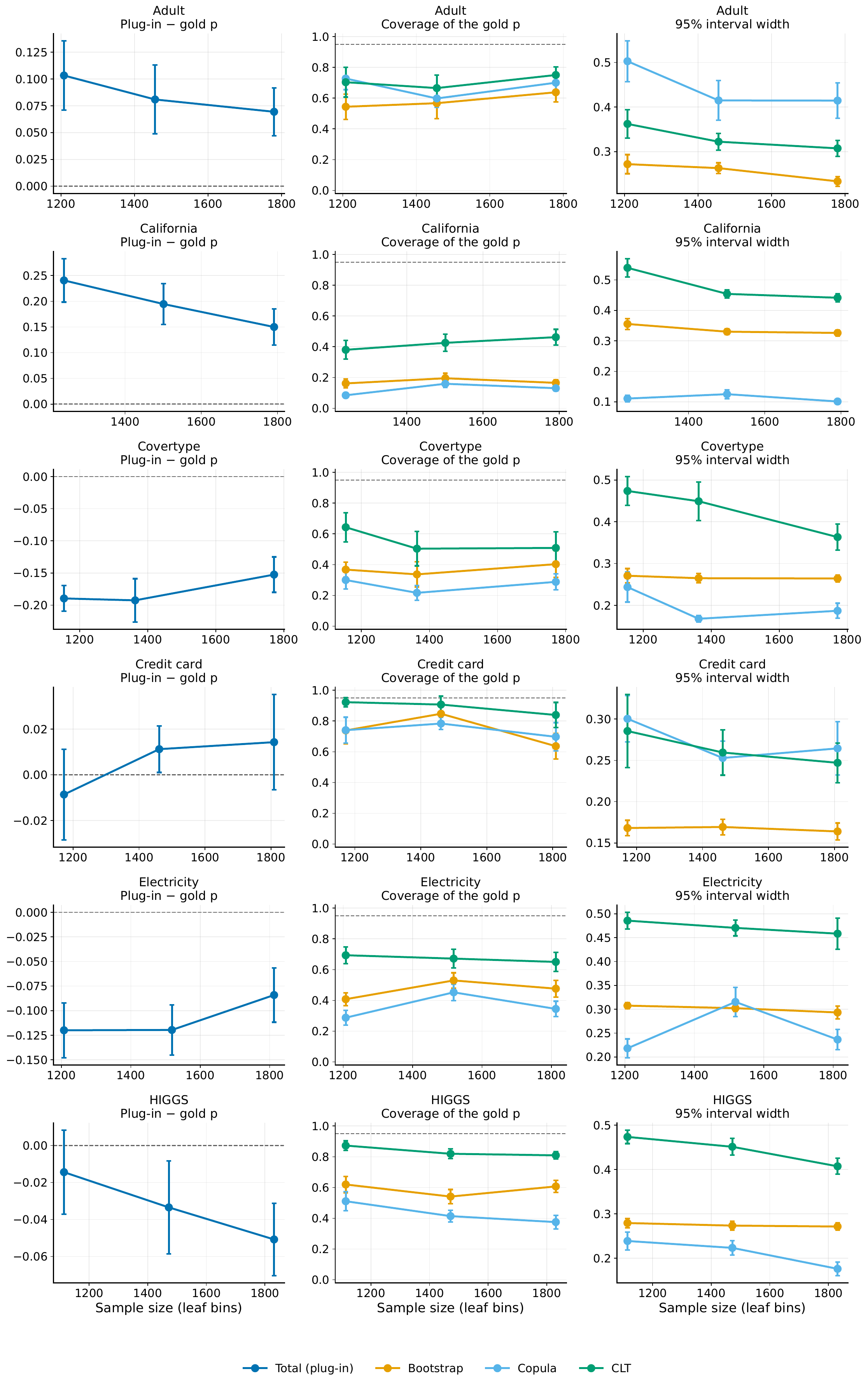}
    \caption{\textbf{Posterior contraction for single-step acquisition across sample size on Tabular Benchmarks.}}
    \label{fig:tabular_sanity_check}
\end{figure}

\clearpage
\paragraph{Posterior contraction for single-step acquisition across sample size on Sepsis.}
Across subsampled datasets of $n \in \{500, 1000, 1500\}$, we validate whether the resulting posterior samples for $P(Y \mid S_T)$ contract (reduce interval width as sample size increases).

\begin{figure}[!ht]
    \centering
    \includegraphics[width=0.9\linewidth]{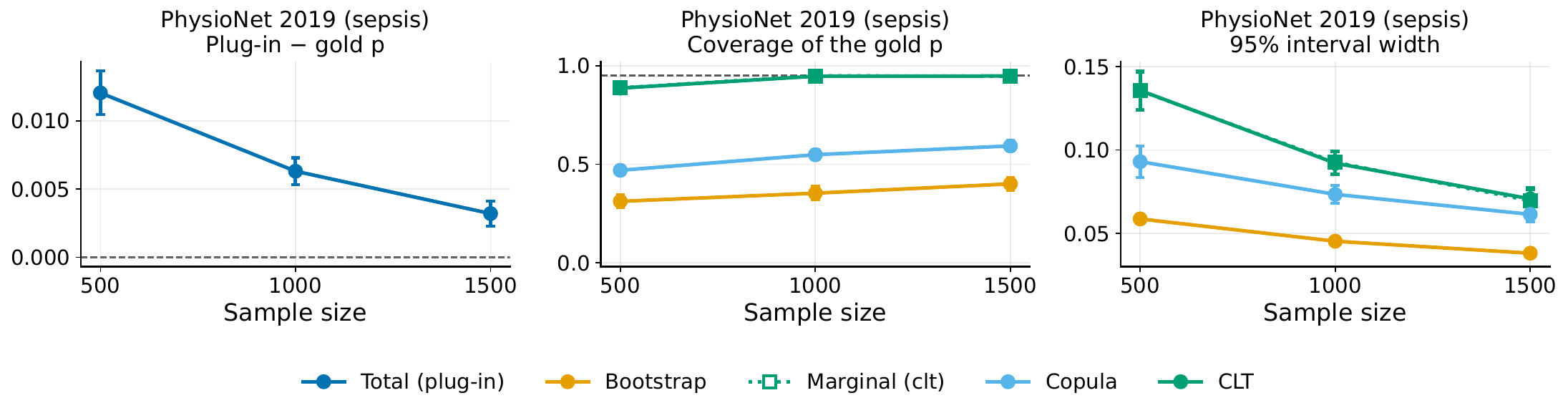}
    \caption{\textbf{Posterior contraction for single-step acquistiion across sample size on Sepsis.} }
    \label{fig:ehr_sanity_check}
\end{figure}

We observe significant heterogeneity in model behavior across the tabular benchmarks. For example, in the \textbf{Credit Card} dataset (\cref{fig:tabular_sanity_check}), the plug-in PPD estimates at $P_n$ are already very close to the reference $P_N$ and exhibit high coverage across all posteriors, meaning the epistemic uncertainty remains largely unchanged. The \textbf{Sepsis} (\cref{fig:ehr_sanity_check}) and \textbf{Adult} task shows stable behavior where $P_n$ converges to the reference and the posterior intervals across all methods become narrow as more data is observed. The copula-based martingale posterior underestimates uncertainty and the interval width is severly underestimated for some datasets such as \textbf{California} and \textbf{Higgs}.
\end{document}